\documentclass[lettersize,journal]{IEEEtran}
\usepackage{amsmath,amsfonts}

\usepackage{array}
\usepackage[caption=false,font=footnotesize,labelfont=rm,textfont=rm]{subfig}
\usepackage{textcomp}
\usepackage{stfloats}
\usepackage{url}
\usepackage{verbatim}
\usepackage{graphicx}
\usepackage{cite}
\usepackage[linesnumbered, ruled]{algorithm2e}
\usepackage{epsfig}
\usepackage{amsmath}
\usepackage{amssymb}
\usepackage{algpseudocode}
\usepackage{xcolor}
\usepackage{amsfonts} 
\usepackage{mathtools}
\usepackage{array,multirow}
\usepackage{bigstrut,bigdelim}
\usepackage{blindtext}
\usepackage{booktabs}
\usepackage{diagbox}
\usepackage{bm}
\usepackage{enumerate}
\usepackage{boldline}
\usepackage{makecell}
\usepackage{bbding} 
\usepackage{verbatim}
\usepackage{soul}
\usepackage{setspace}

\begin{document}

\title{Dual Clustering Co-teaching with Consistent Sample Mining for Unsupervised Person Re-Identification}

\author{
	Zeqi Chen$^{1}$\quad Zhichao Cui$^{2}$ 
	\quad Chi Zhang$^{1}$\quad Jiahuan Zhou$^{3}$ \quad Yuehu Liu$^{1}$\\

	$^{1}$ Xi'an Jiaotong University \quad $^{2}$ Chang'an University \quad $^{3}$ Peking University  \\
	\thanks{This work has been submitted to the IEEE for possible publication. Copyright may be transferred without notice, after which this version may no longer be accessible.}
}



\maketitle

\begin{abstract}
	
	In unsupervised person Re-ID, peer-teaching strategy leveraging two networks to facilitate training has been proven to be an effective method to deal with the pseudo label noise. However, training two networks with a set of noisy pseudo labels reduces the complementarity of the two networks and results in label noise accumulation. To handle this issue, this paper proposes a novel Dual Clustering Co-teaching (DCCT) approach. DCCT mainly exploits the features extracted by two networks to generate two sets of pseudo labels separately by clustering with different parameters. Each network is trained with the pseudo labels generated by its peer network, which can increase the complementarity of the two networks to reduce the impact of noises. Furthermore, we propose dual clustering with dynamic parameters (DCDP) to make the network adaptive and robust to dynamically changing clustering parameters. Moreover, Consistent Sample Mining (CSM) is proposed to find the samples with unchanged pseudo labels during training for potential noisy sample removal. Extensive experiments demonstrate the effectiveness of the proposed method, which outperforms the state-of-the-art unsupervised person Re-ID methods by a considerable margin and surpasses most methods utilizing camera information.

\end{abstract}


\section{Introduction}

\IEEEPARstart{P}{erson} re-identification (Re-ID) aims to retrieve the images of the same person captured by different cameras~\cite{zheng2016person}. Although supervised person Re-ID~\cite{tian2019person,2020feature,Park_Ham_2020,zhou2020online,2021large} has achieved excellent accuracy on publicly available datasets~\cite{2015market1501,2018PTGAN_msmt17,2019personx}, the requirement of tremendous manual annotation limits their practicality in the real world. To tackle this issue, unsupervised person Re-ID methods \cite{2019BUC,2020HCT,2021ICE,2022PPLR,2022ISE} without any labeled data have been extensively studied.

The mainstream unsupervised methods are clustering-based \cite{2019BUC,2020HCT,2020SPCL,2022PPLR,2022ISE}, which are mainly divided into two stages: (1) generating pseudo labels by clustering; (2) training the network with pseudo labels. Although those methods achieve excellent performance, their generated pseudo labels are inevitably noisy. On the one hand, person images with different identities may have similar appearance, viewpoint, pose, and illumination. Due to subtle differences, they may be clustered into a cluster by clustering algorithms. On the other hand, the images of a person may have occlusion, different resolution, and motion blur. They may be clustered into different clusters due to their distinct differences. Training with noisy pseudo labels hinders the model's performance.

To mitigate the influence of such noisy pseudo labels, a peer-teaching strategy \cite{2018coteaching,2019comining,2017meanteachers,2020ACT,2020MMT} is employed, which leverages the difference and complementarity of the two networks to filter different noises through cooperative training of the two networks. ACT \cite{2020ACT} trains two networks in an asymmetric manner to enhance the complementarity of the two networks. One network is trained with pure samples, while the other is trained with diverse samples. To enhance the output independence of the two networks, MMT \cite{2020MMT} utilizes the outputs of the network's temporally average model \cite{2017meanteachers} as soft pseudo labels to train its peer network. However, both ACT and MMT adopt only one set of noisy pseudo labels to train the two networks, resulting in the accumulation and propagation of pseudo label noise during training.

\IEEEpubidadjcol

To overcome the aforementioned shortcomings, we propose a novel Dual Clustering Co-teaching (DCCT) framework to train two networks using two sets of pseudo labels. Training with pseudo labels obtained by different clusterings can increase the differences and complementarity of the two networks, thereby reducing the effect of noises and improving the final performance. Specifically, we propose dual clustering with dynamic parameters (DCDP) to obtain different clustering parameters at each epoch. Then, the features extracted by the temporally average models ($Mean \ Nets$) of two networks are clustered to generate two sets of pseudo labels. And two memory banks are initialized according to the clustering results, as shown in Fig. \ref{Figure_framework}\subref{Figure_framework_a}. Then we adopt the pseudo labels generated by one network to train its peer network, as shown in Fig. \ref{Figure_framework}\subref{Figure_framework_b}. In addition, we propose consistent sample mining (CSM) in each mini-batch to discard potential noisy samples with incorrect pseudo labels, which improves the network's performance.

The main contributions of this paper can be summarized as threefold:
\begin{itemize}
	\item We design a novel peer-teaching framework called Dual Clustering Co-teaching (DCCT), which employs dual clustering with dynamic parameters (DCDP) to generate two sets of pseudo labels. Training with different pseudo labels can enhance the differences and complementarity of the two networks and improve their final performance. The proposed DCDP is so flexible to be effective on multiple clustering algorithms.
	\item We also propose consistent sample mining (CSM) to discard the samples whose pseudo labels are inconsistent during each training epoch. The discarded inconsistent samples are potential noisy samples that may hinder network training.
	\item Extensive experiments on three large-scale datasets (Market-1501~\cite{2015market1501}, MSMT17~\cite{2018PTGAN_msmt17}, and PersonX~\cite{2019personx}) demonstrate that our method outperforms the fully unsupervised state-of-the-art methods by a large margin, even surpasses most UDA methods and methods utilizing camera information.
\end{itemize}

\section{Related Works}
\subsection{Unsupervised Person Re-ID}
Unsupervised person Re-ID methods are mainly divided into unsupervised domain adaptive (UDA) methods and unsupervised learning (USL) methods.

\subsubsection{UDA Person Re-ID}
UDA methods generally pre-train a model using labeled data on the source domain and transfer the learned knowledge from the source domain to the unlabeled target domain. Recent studies in UDA method for person Re-ID can mainly group into clustering-based adaptation \cite{2018PUL,2020Theory,2020adcluster,2022adadc} and cross-domain translation \cite{2020DAAL,2018PTGAN_msmt17,2018SPGAN,2019ECN,2020ECN++,2022triple}. 

The clustering-based adaptation method aims to leverage clustering to generate pseudo labels for unlabeled data on the target domain. Fan $et \ al.$ \cite{2018PUL} utilize the pseudo labels generated by $k$-means \cite{1982kmeans} to fine-tune the model. Song $et \ al.$ \cite{2020Theory} adopt DBSCAN \cite{1996DBSCAN} to generate pseudo labels, and the number of clusters is determined by the density of features. The AD-cluster \cite{2020adcluster} leverages iterative density-based clustering to generate pseudo labels. It learns an image generator to augment the training samples to enforce the discrimination ability of Re-ID models. To avoid overfitting to noisy pseudo labels, AdaDC \cite{2022adadc} adaptively and alternately utilizes different clustering methods. Although the clustering-based method has been proven effective and achieves state-of-the-art performance, due to the existence of some indistinguishable persons with similar appearance, the pseudo labels assigned by the clustering method will be inevitably noisy, which will seriously hinder the training of the network.

The cross-domain translation is another approach that learns domain-invariant features from source-domain images. Generative Adversarial Network (GAN) is one of the main representatives of this type of method. PTGAN \cite{2018PTGAN_msmt17} and SPGAN \cite{2018SPGAN} utilize the images of the source domain to generate the transferred images that have the same style as the target domain images. However, the quality of the generated images restricts the performance of such methods. DAAL \cite{2020DAAL} separate the feature map into the domain-shared feature map and the domain-specific feature map simultaneously. The former is transferred from the source domain to the target domain to facilitate the Re-ID task. ECN \cite{2019ECN} and ECN++ \cite{2020ECN++} adopt a feature memory to learn exemplar-invariance, camera-invariance, and neighborhood-invariance. HCN \cite{2022HCN} proposes a heterogeneous convolutional network, which leverages CNN and GCN to learn the appearance and correlation information of person images. TAL-MIRN \cite{2022triple} leverages triple adversarial learning and multi-view imaginative reasoning to improve the generalization ability of the Re-ID model from the source domain to the target domain. Although these UDA methods perform well under the cross-domain scenario, the requirement of tremendous manually annotation largely limits their usage in practice. In addition, UDA methods rely on the transferable knowledge learned from the source domain, but the discriminative information of the target domain may not be fully explored.

\subsubsection{USL Person Re-ID}
USL methods do not require any labeled data. In recent years, clustering-based methods \cite{2019BUC,2020HCT,2021IICS} have become the mainstream of USL methods. BUC \cite{2019BUC} presents bottom-up clustering to generate pseudo labels, and a diversity regularization is employed to control the number of samples in each cluster. However, only one bottom-up clustering is performed in the entire training process, and incorrectly merged samples in the previous merging steps will always affect the subsequent training process. HCT \cite{2020HCT} adopts hierarchical clustering to generate pseudo labels and employs batch hard triplet loss \cite{2017hardbatch} to facilitate training. TSSL \cite{2020TSSL} designs a unified formulation to consider tracklet frame coherence, tracklet neighbourhood compactness, and tracklet cluster structure. In order to improve the generation quality of pseudo labels, IICS \cite{2021IICS} decomposes the sample similarity computation into two stages: intra-camera and inter-camera computation. PPLR \cite{2022PPLR} exploits the complementary relationship between global and local features to reduce pseudo label noise. To reduce ``sub and mixed" clustering errors, ISE \cite{2022ISE} generates support samples around cluster boundaries to associate the same identity samples.

Some studies address unsupervised person Re-ID without using clustering. SSL \cite{2020SSL} explores the similarity between unlabeled images via softened similarity learning. And a cross-camera encouragement term is proposed to boost softened similarity learning. MMCL \cite{2020MMCL} employs the multi-label classification method to tackle unsupervised person Re-ID and proposes a memory-based multi-label classification loss to promote training. Although these methods have achieved satisfactory performance, there is still a gap between them and clustering-based methods.

In the latest researches, some contrastive learning based methods have achieved remarkable performances. SpCL \cite{2020SPCL} stores the features of all instances in hybrid memory and optimizes the encoder with a unified contrastive loss. Cluster-Contrast \cite{2021CC} stores features and computes contrastive loss at the cluster level. CAP \cite{2021CAP} designs both intra-camera and inter-camera contrastive learning to boost training. ICE \cite{2021ICE} employs inter-instance pairwise similarity scores to promote contrastive learning. However, the inevitable pseudo label noise limits the performance of these methods.

\subsection{Learning with Noisy Labels}
In recent years, training networks on noisy or unlabeled data has been widely studied, which can be classified into four categories: estimating the noise transition matrix \cite{2017transition_matrix1,2017transition_matrix2}, designing the robust loss function \cite{2017robustloss_1,2018robustloss_2}, correcting the noisy labels \cite{2018cleannet,2019deepselflearning} and utilizing peer-teaching strategy \cite{2019comining,2018coteaching,2020MMT}.

This paper focuses on leveraging the peer-teaching strategy method to alleviate label noise. Co-teaching \cite{2018coteaching} trains two networks, and each network selects the samples with small losses to train its peer network. Inspired by Co-teaching, Co-mining \cite{2019comining} trains two networks for face recognition tasks, and the clean samples in each mini-batch are re-weighted. Mean teachers \cite{2017meanteachers} average model weights to deal with large datasets and achieve better performance than averaging label predictions. Drawing inspiration from Co-teaching, ACT \cite{2020ACT} trains two networks in an asymmetric way to tackle unsupervised person Re-ID. However, one of the networks is only trained with clean samples, which limits its generalization capacity. MMT \cite{2020MMT} employs the peer-teaching strategy on unsupervised person Re-ID, and proposes to utilize the temporally average model \cite{2017meanteachers} to generate pseudo labels and soft pseudo labels to avoid training error amplification. However, it leverages noisy pseudo labels to train two networks simultaneously, which results in noise accumulation and affects the performance of the model.

\begin{figure*}
	\centering
	\begin{minipage}{0.44\linewidth}
		\centering		
		\subfloat[Dual clustering with dynamic parameters (DCDP) for pseudo label generation and memory initialization.]{
			\includegraphics[width=0.97\textwidth,trim=2 685 270 4,clip]{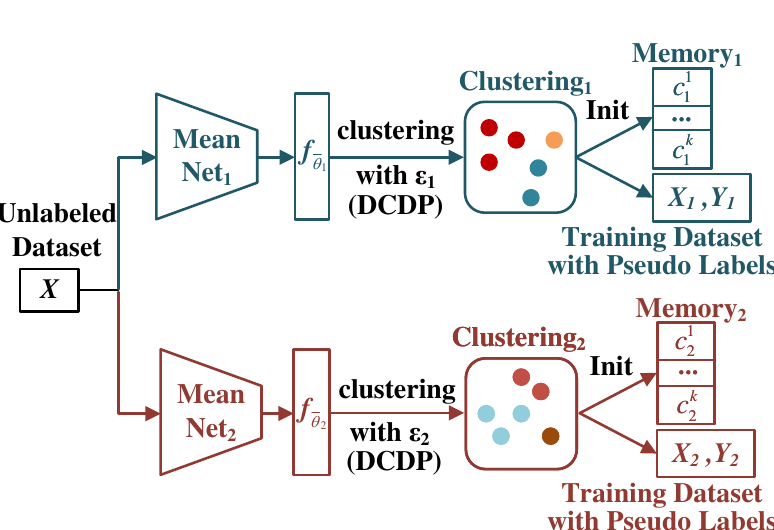}
			\label{Figure_framework_a}		
		}\\
		\setcounter{subfigure}{2}
		\vspace{-3pt}
		\subfloat[One of the Mean Nets is employed for inference.]{
			\includegraphics[width=0.97\textwidth,trim=2 609 270 161,clip]{./Figure/Figure_framework_abc.pdf}
			\label{Figure_framework_c}
		}
	\end{minipage}
	\hspace{5pt}
	\begin{minipage}{0.53\linewidth}
		\vspace{4pt}
		\centering
		\setcounter{subfigure}{1}
		\subfloat[Consistent sample mining (CSM) and co-teaching.]{
			\includegraphics[width=0.97\textwidth,trim=230 600 1 4,clip]{./Figure/Figure_framework_abc.pdf}
			\label{Figure_framework_b}
		}
		
	\end{minipage}
	
	\caption{The framework of proposed Dual Clustering Co-teaching (DCCT) approach. In order to show the co-teaching process of the two networks more clearly, $Net_1$ and its results are shown in blue, while $Net_2$ and its results are shown in red. (a) The features extracted by $Mean \ Net_1$ and $Mean \ Net_2$ are clustered with different parameters $\varepsilon_1$ and $\varepsilon_2$ at each epoch to generate two sets of pseudo labels and initialize two memory banks. $\varepsilon_1$ and $\varepsilon_2$ change dynamically during training, so we call it dual clustering with dynamic parameters (DCDP). (b) Consistent sample mining (CSM) is performed in each iteration. Specifically, $Mean \ Net_1$ and $Memory_1$ are employed to mine consistent samples $X_1^*$ from training dataset $X_1$. Then $X_1^*$ is adopted to train $Net_2$. (The training of $Net_1$ is similar.) The contrastive loss shown in Eq. \ref{loss} is used for training. (c) Since the performance of $Mean \ Net$ is better than that of $Net$, one of the $Mean \ Net$ with better performance is employed for inference. More details are narrated in Algorithm \ref{alg_Framwork}.}
	\label{Figure_framework}
\end{figure*}

\section{Dual Clustering Co-teaching (DCCT)}

Inspired by previous peer-teaching strategy methods\cite{2020ACT,2020MMT}, we develop a novel Dual Clustering Co-teaching (DCCT) framework to train two networks using two sets of pseudo labels. To increase the difference and independence of the two networks, we follow MMT \cite{2020MMT} to employ the temporally average models \cite{2017meanteachers} for our method.

\subsection{The Framework of DCCT}

In order to better illustrate the workflow of our method, the two networks are called $Net_1$ and $Net_2$ for short, and their temporally average models are called $Mean \ Net_1$ and $Mean \ Net_2$. As shown in Fig. \ref{Figure_framework}, our method mainly contains two stages: (a) pseudo label generation and memory initialization; (b) co-teaching of the two networks. 

\textbf{(a) In the stage of pseudo label generation and memory initialization}, we adopt $Mean \ Net_1$ and $Mean \ Net_2$ to extract features from the unlabeled dataset $X$. Then, different clustering parameters are calculated according to the proposed dual clustering with dynamic parameters (DCDP). After that, the features extracted by $Mean \ Net_1$ and $Mean \ Net_2$ are clustered with different parameters to generate two sets of pseudo labels. And some outliers in $X$ may be discarded according to the clustering results. Then, we get two relatively clean datasets and their pseudo labels: training dataset $X_1$ with pseudo labels $Y_1$ and training dataset $X_2$ with pseudo labels $Y_2$. At the same time, the two clustering results are exploited to initialize the memory bank $Memory_1$ and $Memory_2$, as shown in Fig. \ref{Figure_framework}\subref{Figure_framework_a}.

\textbf{(b) In the stage of co-teaching}, we perform consistent sample mining (CSM) in each iteration. Concretely, we leverage $Mean \ Net_1$ and $Memory_1$ to mine consistent sample set $X_1^*$ from the training dataset $X_1$, and $X_1^*$ is employed to train $Net_2$. Similarly, the consistent sample set $X_2^*$ is employed to train $Net_1$, as shown in Fig. \ref{Figure_framework}\subref{Figure_framework_b}. $Net_1$ and $Net_2$ are trained by the contrastive loss shown in Eq. \ref{loss}. $Memory_1$ and $Memory_2$ are updated by the momentum update strategy shown in Eq. \ref{memory_update}, while $Mean \ Net_1$ and $Mean \ Net_2$ are updated by Eq. \ref{average_network_update}.

Compared with previous methods, we mainly made two contributions: (1) In stage (a), we proposed dual clustering with dynamic parameters (DCDP) to promote network training by generating two sets of pseudo labels (Sec. \ref{Detail_DCDP}). (2) In stage (b), we proposed consistent sample mining (CSM) to remove potential noise samples (Sec. \ref{Detail_CSM}). More details of DCCT's procedure are narrated in Algorithm \ref{alg_Framwork}.

\subsection{Pseudo Label Generation and Memory Initialization}
\label{Detail_DCDP}

\subsubsection{Dual Clustering with Dynamic Parameters (DCDP) for Pseudo Label Generation}
Training two networks with one set of pseudo labels suffers from three limitations. (1) Utilizing the same data and supervision to train two networks makes them too similar and lose their complementarity and differences. (2) Leveraging the same noisy pseudo labels to train two networks results in error accumulation and propagation. (3) Using two features to generate a set of pseudo labels may lose some information because it is not easy to find a reasonable and effective way to fuse the features extracted by the two networks.

To handle the aforementioned issues, we propose to train two networks separately with different pseudo labels generated by different clusterings, which can increase their differences and complementarity. Therefore, the samples that cannot be discriminated well by one network may be well discriminated by its peer network, so that the two networks can filter different noises and better collaborative teaching. Furthermore, to make the network adaptive and robust to different clustering parameters, the clustering parameters of the same network can also be dynamically changed to enhance the network's generalization ability. Based on the above considerations, we proposed dual clustering with dynamic parameters (DCDP) for pseudo label generation. Although the proposed DCDP can be combined with multiple clustering algorithms (such as DBSCAN \cite{1996DBSCAN}, $k$-means \cite{1982kmeans}, and InfoMap \cite{2008infomap}), the DBSCAN is exploited in our framework thanks to its superior ability. The effect of DCDP on other clustering algorithms is demonstrated in Sec. \ref{DCDP_on_other}.

\textbf{The details of DCDP.} The maximum distance between two samples is the most crucial parameter in DBSCAN \cite{1996DBSCAN}, which is adopted as the dynamic parameter in DCDP. Given the initial value $\varepsilon$ and the increment size $\Delta{\varepsilon}$ of the maximum distance, the maximum distance $\varepsilon_1$ and $\varepsilon_2$ of $clustering_1$ and $clustering_2$ will vary in the range of [$\varepsilon - \Delta{\varepsilon}$, $\varepsilon + \Delta{\varepsilon}$]. The number of training epochs is recorded as $E$, then the $\varepsilon_1^i$ and $\varepsilon_2^i$ at the $i$-th epoch can be obtained from Eq. \ref{calculation_of_eps}.

\begin{equation}
	\begin{split}
		\varepsilon_1^i=\left\{
		\begin{aligned}
			\varepsilon + \frac{2\Delta{\varepsilon}}{E}i, \quad 0 \leq i < \frac{E}{2}\\
			\varepsilon + 2\Delta{\varepsilon} - \frac{2\Delta{\varepsilon}}{E}i, \quad \frac{E}{2} \leq i \leq E \\
		\end{aligned}\quad,
		\right.
		\\
		\varepsilon_2^i=\left\{
		\begin{aligned}
			\varepsilon - \frac{2\Delta{\varepsilon}}{E}i, \quad 0 \leq i < \frac{E}{2}\\
			\varepsilon - 2\Delta{\varepsilon} + \frac{2\Delta{\varepsilon}}{E}i, \quad \frac{E}{2} \leq i \leq E \\
		\end{aligned}\quad.
		\right.
	\end{split}
	\label{calculation_of_eps}
\end{equation}

At the start of each epoch, $Mean \ Net_1$ and $Mean \ Net_2$ are employed to extract the features of unlabeled dataset $X$. Then, the extracted features $\boldsymbol{f}_{\bar\theta_1}$ and $\boldsymbol{f}_{\bar\theta_2}$ are utilized for clustering with parameters $\varepsilon_1$ and $\varepsilon_2$. Since DBSCAN removes outliers, we obtain two relatively clean training datasets $X_1$ and $X_2$, and their pseudo labels $Y_1$ and $Y_2$.

\subsubsection{Memory Initialization}

As shown in Fig. \ref{Figure_framework}\subref{Figure_framework_a}, $clustering_1$ is employed to initialize the memory bank $Memory_1$. Following Cluster-Contrast \cite{2021CC}, the mean feature vectors of each cluster are adopted to initialize the cluster representations $\{\boldsymbol{c}_1^1,...,\boldsymbol{c}_1^K\}$, where $\boldsymbol{c}_{1}^{k}$ denotes the $k$-th cluster representation of $Memory_1$ and $K$ is the cluster number. So $Memory_1$ is initialized by:

\begin{equation}
	\begin{aligned}
		\boldsymbol{c}_{1}^{k}=\frac{1}{|C_{1}^{k}|}\sum_{\boldsymbol{x}_1^j \in C_{1}^{k}}\boldsymbol{f}_{\bar\theta_1}(\boldsymbol{x}_1^j),
	\end{aligned}
	\label{memory-initialization}
\end{equation}
where $C_{1}^{k}$ denotes the $k$-th cluster of $clustering_1$ and $|\cdot|$ indicates the number of instances per cluster. $\boldsymbol{x}_1^j$ denotes the $j$-th samples in $X_1$, and $\boldsymbol{f}_{\bar\theta_1}(\cdot)$ denotes the features extracted by $Mean \ Net_1$. The initialization of the memory bank $Memory_2$ is similar to that of $Memory_1$.

\begin{algorithm}[t]
	\caption{Procedure of the DCCT.} 
	\label{alg_Framwork} 
	
	\KwIn {
		unlabeled dataset $X$; 
		ImageNet pre-trained ResNet-50 $\boldsymbol{\theta}$; 
		maximum distance $\varepsilon$ and its increment size $\Delta{\varepsilon}$ for Eq. 1; 	
		threshold $\gamma$ for clustering quality; 	
		temperature $\tau$ for Eq. 4; 	
		momentum $\beta$ for Eq. 5; 	
		momentum $\alpha$ for Eq. 6; 	
		maximal epoch $E$; 	
		maximal iteration $I$.
	}
	\KwOut {Best $Mean \ Net$ $\boldsymbol{\bar{\theta}}^*$ after training.}

	Initialize: $Net_1$ $\boldsymbol{\theta}_1 \gets \boldsymbol{\theta}$, $Net_2$ $\boldsymbol{\theta}_2 \gets \boldsymbol{\theta}$, $Mean \ Net_1$ $\boldsymbol{\bar{\theta}}_1 \gets \boldsymbol{\theta}_1$, $Mean \ Net_2$ $\boldsymbol{\bar{\theta}}_2 \gets \boldsymbol{\theta}_2$;
	
	\For{$epoch = 1 \ \KwTo \ E$}
	{
		Extract feature $\boldsymbol{f}_{\bar\theta_1}$ and $\boldsymbol{f}_{\bar\theta_2}$ from $X$ by $\boldsymbol{\bar{\theta}}_1$ and $\boldsymbol{\bar{\theta}}_2$;
		
		Calculate parameters $\varepsilon_1$ and $\varepsilon_2$ for DCDP by Eq. \ref{calculation_of_eps};
		
		Perform clustering on $\boldsymbol{f}_{\bar\theta_1}$ and $\boldsymbol{f}_{\bar\theta_2}$;
		
		Generate training dataset $X_1$, $X_2$ and their pseudo labels $Y_1$, $Y_2$ by two clustering results;
		
		Initialize $Memory_1$ and $Memory_2$ by Eq. \ref{memory-initialization};
		
		Calculate $DBI_1$ and $DBI_2$ of two clusterings;
		
		\For{$iter = 1 \ \KwTo \ I$}
		{
			\eIf{$min(DBI_1,DBI_2) < \gamma$}
			{
				Perform CSM by Eq. \ref{CSM-equation} to obtain $X_1^*$ and $X_2^*$;
			}{
				$X_1^*=X_1$, $X_2^*=X_2$;
			}

			Train $\boldsymbol{\theta}_1$ ($\boldsymbol{\theta}_2$) using $X_2^*$ ($X_1^*$) and loss function in Eq. \ref{loss};
			
			Update $Memory_1$ and $Memory_2$ by Eq. \ref{memory_update};
			
			Update $\boldsymbol{\bar{\theta}}_1$ and $\boldsymbol{\bar{\theta}}_2$ by Eq. \ref{average_network_update};
		}
	}
\end{algorithm}

\subsection{Consistent Sample Mining and Co-teaching}
\label{Detail_CSM}

\subsubsection{Consistent Sample Mining (CSM)}
Directly using noisy pseudo labels to train the network will reduce its final performance. Therefore, we propose consistent sample mining (CSM) to extract consistent samples and remove potential noise samples. Since the trainings of $Net_1$ and $Net_2$ are similar, we only introduce the training details of $Net_2$ and CSM details of $Mean \ Net_1$ and $Memory_1$.

\textbf{Inconsistency of pseudo labels.} For any sample $\boldsymbol{x}_1^j$ in training datasets $X_1$, its feature extracted by $Mean \ Net_1$ is denoted as $\boldsymbol{f}_{\bar\theta_1}(\boldsymbol{x}_1^j)$. And $y_1^j$ represents the pseudo label of $\boldsymbol{x}_1^j$. At the beginning of each epoch, the features $\boldsymbol{f}_{\bar\theta_1}$ extracted by $Mean \ Net_1$ are employed for clustering, and the clustering results are utilized to initialize $Memory_1$ and generate pseudo labels (see Fig. \ref{Figure_framework}\subref{Figure_framework_a}). At this time, calculating the similarity between $\boldsymbol{f}_{\bar\theta_1}(\boldsymbol{x}_1^j)$ and each cluster representation $\boldsymbol{c}_{1}^{k}$ stored in $Memory_1$, $\boldsymbol{f}_{\bar\theta_1}(\boldsymbol{x}_1^j)$ will be most similar to the cluster indicated by the pseudo label $y_1^j$.

In each iteration, the parameters of $Mean \ Net_1$ are updated by parameters of $Net_1$ with momentum $\alpha$ (see Fig. \ref{Figure_framework}\subref{Figure_framework_b} and Eq. \ref{average_network_update}), and each clustering representation $\boldsymbol{c}_{1}^{k}$ stored in $Memory_1$ is updated by the features extracted by $Net_2$ with momentum $\beta$ (see Fig. \ref{Figure_framework}\subref{Figure_framework_b} and Eq. \ref{memory_update}), while the pseudo labels are not updated synchronously. At this time, calculating the similarity between each $\boldsymbol{f}_{\bar\theta_1}(\boldsymbol{x}_1^j)$ and each $\boldsymbol{c}_{1}^{k}$, some samples may be most similar to the clustering representations that are inconsistent with their pseudo labels. 

\textbf{Definition of consistent samples.} 
In each iteration, we calculate the cosine similarity between each sample's feature $\boldsymbol{f}_{\bar\theta_1}(\boldsymbol{x}_1^j)$ and each clustering representation $\boldsymbol{c}_{1}^{k}$. Then we obtain the most similar cluster $k^*$ of each sample $\boldsymbol{x}_1^j$ by:

\begin{equation}
	\begin{aligned}
		k^* = \underset{k \in \{1,2,...,K\}}{\arg\max} sim(\boldsymbol{f}_{\bar\theta_1}(\boldsymbol{x}_1^j), \boldsymbol{c}_{1}^{k}),
	\end{aligned}
	\label{CSM-equation}
\end{equation}
where $sim()$ denotes the cosine similarity between two vectors. When $k^*$ is consistent with the pseudo label of $\boldsymbol{x}_1^j$, the sample $\boldsymbol{x}_1^j$ is considered consistent. Otherwise, the sample is considered inconsistent, and its pseudo label is changed to -1.

We argue that inconsistent samples hamper network training, while only consistent samples should be employed for training. The number of consistent samples increases with the gradual convergence of the network (see Fig. \ref{CSM_Figure}\subref{CSM_Figure_a}). Eventually, it tends to exploit all samples for training, which is in accordance with the concept of self-paced learning \cite{2010self}.

\textbf{Clustering quality evaluation for CSM.}
However, in the early stage of training, the clustering quality may be poor. Using CSM at this time, the number of consistent samples selected in each iteration may be too small, which impairs network training. Therefore, we propose to employ the Davies-Bouldin index (DBI) \cite{1979DBI} to measure the clustering quality. DBI is an internal clustering evaluation scheme without the demand for ground truth. The lower bound of the DBI is 0, and a lower DBI value means a better clustering quality. Therefore, we set a \textbf{threshold $\gamma$} to judge whether the clustering quality is good. When DBI is less than $\gamma$, the clustering quality is considered good enough to mine consistent samples. We analyze how the parameter $\gamma$ affects the network performance in Sec. \ref{Sec_parameter_analysis}.

\subsubsection{Loss Function}

For any query instances, its features extracted by $Net_2$ are recorded as $\boldsymbol{q}$, which is compared to all the cluster representations $\{\boldsymbol{c}_1^1,...,\boldsymbol{c}_1^K\}$ stored in $Memory_1$ using the following InfoNCE loss:

\begin{equation}
	\begin{aligned}
		L_{\boldsymbol{q}} = -\log\frac{\exp(\boldsymbol{q} \cdot \boldsymbol{c}_1^+) / \tau}{\sum_{k=1}^{K} \exp(\boldsymbol{q} \cdot \boldsymbol{c}_1^k) / \tau},
	\end{aligned}
	\label{loss}
\end{equation}
where $\boldsymbol{c}_1^+$ is the clustering representation indicated by the pseudo label of the query instance, $\tau$ is a temperature hyper-parameter \cite{2018tau}.

\subsubsection{Memory Updating}

Following Cluster-Contrast \cite{2021CC}, we adopt the momentum update strategy to update $Memory_1$, which is formulated as follows:
\begin{equation}
	\begin{aligned}
		\boldsymbol{c}_1^k \gets \beta \boldsymbol{c}_1^k + (1-\beta)\boldsymbol{q}^k,
	\end{aligned}
	\label{memory_update}
\end{equation}
where $\boldsymbol{q}^k$ is the sample feature extracted by $Net_2$, which has the same identity as the cluster representations $\boldsymbol{c}_1^k$, and $\beta$ is the ensembling momentum to be within the range of $[0,1)$. 

\subsubsection{Temporally Average Models Updating}

Let $\boldsymbol{\theta}^{i}$ and $\boldsymbol{\bar\theta}^{i}$ denote the parameters of a network and the network's temporally average model at iteration $i$, then $\boldsymbol{\bar\theta}^{i+1}$ can be updated as

\begin{equation}
	\begin{aligned}
		\boldsymbol{\bar\theta}^{i+1} = \alpha \boldsymbol{\bar\theta}^{i} + (1-\alpha)\boldsymbol{\theta}^{i},
	\end{aligned}
	\label{average_network_update}
\end{equation}
where $\alpha$ is the ensembling momentum to be within the range of $[0,1)$. The initial parameters of the temporally average model are $\boldsymbol{\bar\theta}^{0} = \boldsymbol{\theta}^{0}.$

\begin{table*}[!t]
	\caption{Comparison of the proposed DCCT and state-of-the-art methods on Market-1501 and MSMT17. The ``Labels" column lists the type of labels used by the method. ``Transfer" denotes that the manually annotated labels from another Re-ID dataset are utilized for training. ``Camera'' means the camera information is employed by the method. ``None'' means that it is a fully unsupervised method. $\dag$ indicates that the results are reproduced by the author of MMT \cite{2020MMT} in the paper SpCL \cite{2020SPCL}. Performances surpassing all competing methods are \textbf{bold}, and the second-best performances are highlighted using \underline{underline}.}
	
	\centering
	\resizebox{1.0\textwidth}{!}{

		\begin{tabular}{c|c|c|cccc|cccc}
			\hline
			\multirow{2}[4]{*}{\textbf{Method}} & \multirow{2}[4]{*}{\textbf{Reference}} & \multirow{2}[4]{*}{\textbf{Labels}} & \multicolumn{4}{c|}{\textbf{Market-1501}} & \multicolumn{4}{c}{\textbf{MSMT17}} \bigstrut\\
			\cline{4-11}      &       &       & mAP   & top-1 & top-5 & top-10 & mAP   & top-1 & top-5 & top-10 \bigstrut\\
			\hline
			PUL \cite{2018PUL} & TOMM’18 & Transfer & 20.5  & 45.5  & 60.7  & 66.7  & -     & -     & -     & - \bigstrut[t]\\
			PTGAN \cite{2018PTGAN_msmt17} & CVPR'2018 & Transfer & -     & -     & -     & -     & 3.3   & 11.8  & -     & 27.4  \\
			SPGAN \cite{2018SPGAN} & CVPR'2018 & Transfer & 22.8  & 51.5  & 70.1  & 76.8  & -     & -     & -     & - \\
			TAL-MIRN \cite{2022triple} & TCSVT’22 & Transfer & 42.9  & 74.6  & 87.6  & -     & 14.2  & 39.0  & 51.5  & - \\
			ACT \cite{2020ACT} & AAAI'2020 & Transfer & 60.6  & 80.5  & -     & -     & -     & -     & -     & - \\
			ECN++ \cite{2020ECN++} & TPAMI’20 & Transfer & 63.8  & 84.1  & 92.8  & 95.4  & 16.0  & 42.5  & 55.9  & 61.5  \\
			MMCL \cite{2020MMCL} & CVPR'2020 & Transfer & 60.4  & 84.4  & 92.8  & 95.0  & 16.2  & 43.6  & 54.3  & 58.9  \\
			AD-Cluster \cite{2020adcluster} & CVPR'2020 & Transfer & 68.3  & 86.7  & 94.4  & 96.5  & -     & -     & -     & - \\
			HCN\cite{2022HCN} & TCSVT’22 & Transfer & 70.5  & 90.7  & -     & -     & 29.9  & 58.7  & -     & - \\
			MMT-kmeans \cite{2020MMT}  & ICLR'2020 & Transfer & 71.2  & 87.7  & 94.9  & 96.9  & 23.3  & 50.1  & 63.9  & 69.8  \\
			MMT-DBSCAN$^{\dag}$ \cite{2020MMT}  & ICLR'2020 & Transfer & 75.6  & 89.3  & 95.8  & 97.5  & 24.0  & 50.1  & 63.5  & 69.3  \\
			SpCL \cite{2020SPCL} & NeurIPS'20 & Transfer & 77.5  & 89.7  & 96.1  & 97.6  & 26.8  & 53.7  & 65.0  & 69.8  \\
			AdaDC \cite{2022adadc} & TCSVT’22 & Transfer & 83.2  & 92.9  & 97.5  & 98.5  & 32.7  & 60.7  & 73.6  & 78.7  \bigstrut[b]\\
			\hline
			SSL \cite{2020SSL} & AAAI'2020 & Camera & 37.8  & 71.7  & 83.8  & 87.4  & -     & -     & -     & - \bigstrut[t]\\
			IICS \cite{2021IICS} & CVPR'2021 & Camera & 72.9  & 89.5  & 95.2  & 97.0  & 26.9  & 56.4  & 68.8  & 73.4  \\
			CAP \cite{2021CAP} & AAAI'2021 & Camera & 79.2  & 91.4  & 96.3  & 97.7  & 36.9  & 67.4  & 78.0  & 81.4  \\
			ICE \cite{2021ICE} & ICCV'2021 & Camera & 82.3  & 93.8  & 97.6  & 98.4  & 38.9  & 70.2  & 80.5  & 84.4  \\
			PPLR \cite{2022PPLR} & CVPR'2022 & Camera & 84.4  & 94.3  & 97.8  & 98.6  & 42.2  & 73.3  & 83.5  & 86.5  \bigstrut[b]\\
			\hline
			BUC \cite{2019BUC} & AAAI'2019 & None  & 38.3  & 66.2  & 79.6  & 84.5  & -     & -     & -     & - \bigstrut[t]\\
			TSSL \cite{2020TSSL} & AAAI'2020 & None  & 43.3  & 71.2  & -     & -     & -     & -     & -     & - \\
			MMCL \cite{2020MMCL} & CVPR'2020 & None  & 45.5  & 80.3  & 89.4  & 92.3  & 11.2  & 35.4  & 44.8  & 49.8  \\
			HCT \cite{2020HCT} & CVPR'2020 & None  & 56.4  & 80.0  & 91.6  & 95.2  & -     & -     & -     & - \\
			SpCL \cite{2020SPCL} & NeurIPS'20 & None  & 73.1  & 88.1  & 95.1  & 97.0  & 19.1  & 42.3  & 55.6  & 61.2  \\
			ICE \cite{2021ICE} & ICCV'2021 & None  & 79.5  & 92.0  & 97.0  & 98.1  & 29.8  & 59.0  & 71.7  & 77.0  \\
			Cluster-Contrast \cite{2021CC} & arXiv'2021 & None  & 82.1  & 92.3  & 96.7  & 97.9  & 27.6  & 56.0  & 66.8  & 71.5  \\
			PPLR \cite{2022PPLR} & CVPR'2022 & None  & 81.5  & 92.8  & 97.1  & 98.1  & 31.4  & 61.1  & 73.4  & 77.8  \\
			ISE \cite{2022ISE} & CVPR'2022 & None  & \underline{85.3} & \underline{94.3} & \textbf{98.0 } & \textbf{98.8 } & \underline{37.0} & \underline{67.6} & \underline{77.5} & \underline{81.0} \\
			DCCT (Ours) & This paper & None  & \textbf{86.3 } & \textbf{94.4 } & \underline{97.7} & \underline{98.5} & \textbf{41.8 } & \textbf{68.7 } & \textbf{79.0 } & \textbf{82.6 } \bigstrut[b]\\
			\hline
		\end{tabular}%

	}
	\hspace{1pt}
	
	\label{tab_sota_1}%
\end{table*}%

\section{Experiments}

\subsection{Datasets and Evaluation Protocols}

\subsubsection{Datasets.}
We evaluate the proposed method on three large-scale datasets - Market-1501~\cite{2015market1501}, MSMT17~\cite{2018PTGAN_msmt17}, and PersonX~\cite{2019personx}.

\textbf{Market-1501} dataset contains 32,668 annotated images of 1,501 identities captured by 6 cameras on a university campus. In Market-1501, 12,936 images of 751 identities are used as the training set, and 19,732 images of 750 identities are utilized as the test set.

\textbf{MSMT17} dataset contains 126,441 annotated images of 4,101 identities captured by 15 cameras. In MSMT17, 32,621 images of 1,041 identities are used as the training set, and 93,820 images of the remaining 3,060 identities are utilized as the test set.

\textbf{PersonX} is a synthetic dataset with manually designed difficulties such as different viewpoints, illumination, occlusions, and backgrounds. It contains 45,792 annotated images of 1,266 identities captured by 6 cameras. In PersonX, 9,840 images of 410 identities are used as the training set, and 35,952 images of the remaining 856 identities are utilized as the test set.

\subsubsection{Evaluating Setting} 
The mean average precision (mAP) \cite{2015mAP} and cumulative matching characteristic (CMC) \cite{2007CMC} curve are employed to evaluate the performance of each method. And the top-1, top-5, and top-10 accuracies are reported to represent the CMC curve.

\begin{table}[b!]
	\caption{Comparison of the proposed DCCT and state-of-the-art methods on PersonX dataset. $\dag$ indicates that the results are reproduced by Cluster-Contrast \cite{2021CC}.}
	
	\centering	
	\resizebox{0.475\textwidth}{!}{

		\begin{tabular}{c|c|cccc}
			\hline
			\multirow{2}[4]{*}{\textbf{Method}} & \multirow{2}[4]{*}{\textbf{Labels}} & \multicolumn{4}{c}{\textbf{PersonX}} \bigstrut\\
			\cline{3-6}      &       & mAP   & top-1 & top-5 & top-10 \bigstrut\\
			\hline
			MMT-DBSCAN$^{\dag}$ \cite{2020MMT}  & Transfer & 78.9  & 90.6  & 96.8  & 98.2  \bigstrut[t]\\
			SPCL$^{\dag}$ \cite{2020SPCL} & Transfer & 78.5  & 91.1  & 97.8  & 99.0  \\
			SPCL$^{\dag}$ \cite{2020SPCL} & None  & 72.3  & 88.1  & 96.6  & 98.3  \\
			Cluster-Contrast \cite{2021CC} & None  & 84.7  & 94.4  & 98.3  & 99.3  \\
			DCCT (Ours) & None  & \textbf{87.6 } & \textbf{95.0 } & \textbf{98.7 } & \textbf{99.4 } \bigstrut[b]\\
			\hline
		\end{tabular}%

	}
	\hspace{1pt}
	
	\label{tab2}%
\end{table}%

\subsection{Implementation Details}
\label{details}
Our method is implemented based on PyTorch on Linux. We employ four NVIDIA RTX 2080Ti GPUs for training and only one GPU for testing. We adopt a pre-trained ResNet-50 \cite{2016ResNet50} on ImageNet \cite{2012imagenet} as the backbone networks to conduct all the experiments. The model is modified following Cluster-Contrast \cite{2021CC}. The $Net_1$ and $Net_2$ are initialized by the ImageNet pre-trained model, and they are updated by the loss function shown in Eq. \ref{loss} with the temperature hyper-parameter $\tau=0.05$. The temporal momentum $\alpha$ in Eq. \ref{average_network_update} is 0.99. The momentum $\beta$ for memory updating in Eq. \ref{memory_update} is 0.1. We set the number of training epochs to be 50, and the number of training iterations is 300. During the training, all images are resized to $320\times128$, and random cropping, flipping as well as random erasing are adopted for data augmentation \cite{2017random}. The batch size is set to 128, which contains 32 identities, and each identity has 4 images. We employ Adam optimizer to train the model with weight decay $5\times10^{-4}$. The initial learning rate is $3.5\times10^{-4}$, which is reduced to 1/10 of its previous value every 20 epochs. Following SpCL \cite{2020SPCL}, we utilize Jaccard distance based on $k$-reciprocal encoding \cite{2017reranking} for clustering, where $k_1$ is set to 30 and $k_2$ is set to 6. In DBSCAN, the minimum number of samples in the neighborhood of the core point is set to 4, and the maximum distance $\varepsilon$ is set to $0.5$, $0.7$, and $0.7$ for Market-1501, MSMT17, and PersonX. The $\Delta{\varepsilon}$ in Eq. \ref{calculation_of_eps} is set to $0.35$, $0.15$ and $0.15$ for Market-1501, MSMT17 and PersonX. The DBI threshold $\gamma$ in Sec. \ref{Detail_CSM} is 1.3.

\subsection{Comparison with state-of-the-art methods}
In Table \ref{tab_sota_1} and Table \ref{tab2}, we compare our method with several state-of-the-art unsupervised methods on three widely-used person Re-ID datasets (Market-1501, MSMT17, and PersonX). We obtain the best performance among all the compared methods with top-1 =$94.4\%$ and mAP =$86.3\%$ on Market-1501. On PersonX, we also achieve the best performance with top-1 =$95.0\%$, and mAP =$87.6\%$. Note that on the above two datasets, our method not only surpasses other fully unsupervised methods by considerable margins, but also outperforms other methods employing camera information and other UDA methods that require a large number of manually labeled data on the source domain. On MSMT17, we outperform the state-of-the-art fully unsupervised method ISE \cite{2022ISE} by considerable margins of $4.8\%$ mAP. We are also better than all UDA methods and most methods that employ camera information.

\begin{table*}[htbp]
	\caption{Ablation study of our proposed dual clustering with dynamic parameters (DCDP) and consistent sample mining (CSM).}
	\centering
	\resizebox{1.0\textwidth}{!}{
		
		\begin{tabular}{c|cccc|cccc|cccc}
			\hline
			\multirow{2}[4]{*}{\textbf{Method}} & \multicolumn{4}{c|}{\textbf{Market-1501}} & \multicolumn{4}{c|}{\textbf{MSMT17}} & \multicolumn{4}{c}{\textbf{PersonX}} \bigstrut\\
			\cline{2-13}      & mAP   & top-1 & top-5 & top-10 & mAP   & top-1 & top-5 & top-10 & mAP   & top-1 & top-5 & top-10 \bigstrut\\
			\hline
			DCCT w/o (DCDP \& CSM) & 82.9  & 92.9  & 97.1  & 98.1  & 38.5  & 65.9  & 76.8  & 81.0  & 86.3  & 94.4  & 98.4  & 99.2  \bigstrut[t]\\
			DCCT w/o CSM & 85.1  & 93.7  & 97.5  & 98.4  & 40.8  & 68.0  & 78.4  & 82.4  & 86.4  & 94.0  & 98.5  & \textbf{99.4 } \\
			DCCT w/o DCDP & 85.1  & 93.6  & 97.3  & 98.0  & 40.3  & 67.6  & 78.7  & 82.4  & 86.7  & 94.4  & 98.4  & \textbf{99.4 } \\
			DCCT  & \textbf{86.3 } & \textbf{94.4 } & \textbf{97.7 } & \textbf{98.5 } & \textbf{41.8 } & \textbf{68.7 } & \textbf{79.0 } & \textbf{82.6 } & \textbf{87.6 } & \textbf{95.0 } & \textbf{98.7 } & \textbf{99.4 } \bigstrut[b]\\
			\hline
		\end{tabular}%

	}
	\hspace{1pt}
	
	\label{tab_ablation}%
\end{table*}%

\begin{figure}[!b]
	\centering
	
	\subfloat[]{
		\includegraphics[width=0.23\textwidth,trim=92 265 125 280,clip]{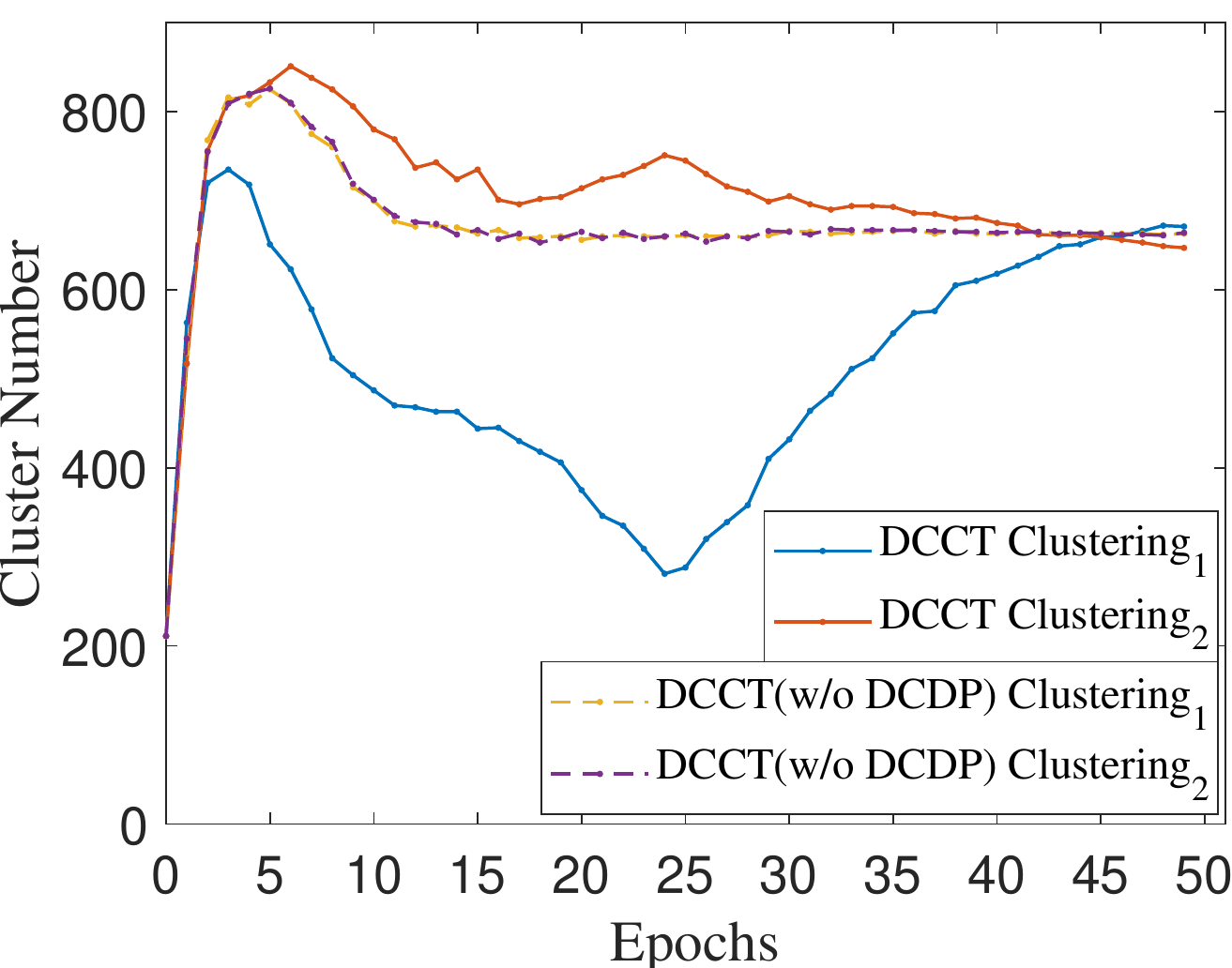}
		\label{DCDP_Figure_a}
	}
	\subfloat[]{
		\includegraphics[width=0.23\textwidth,trim=92 265 125 280,clip]{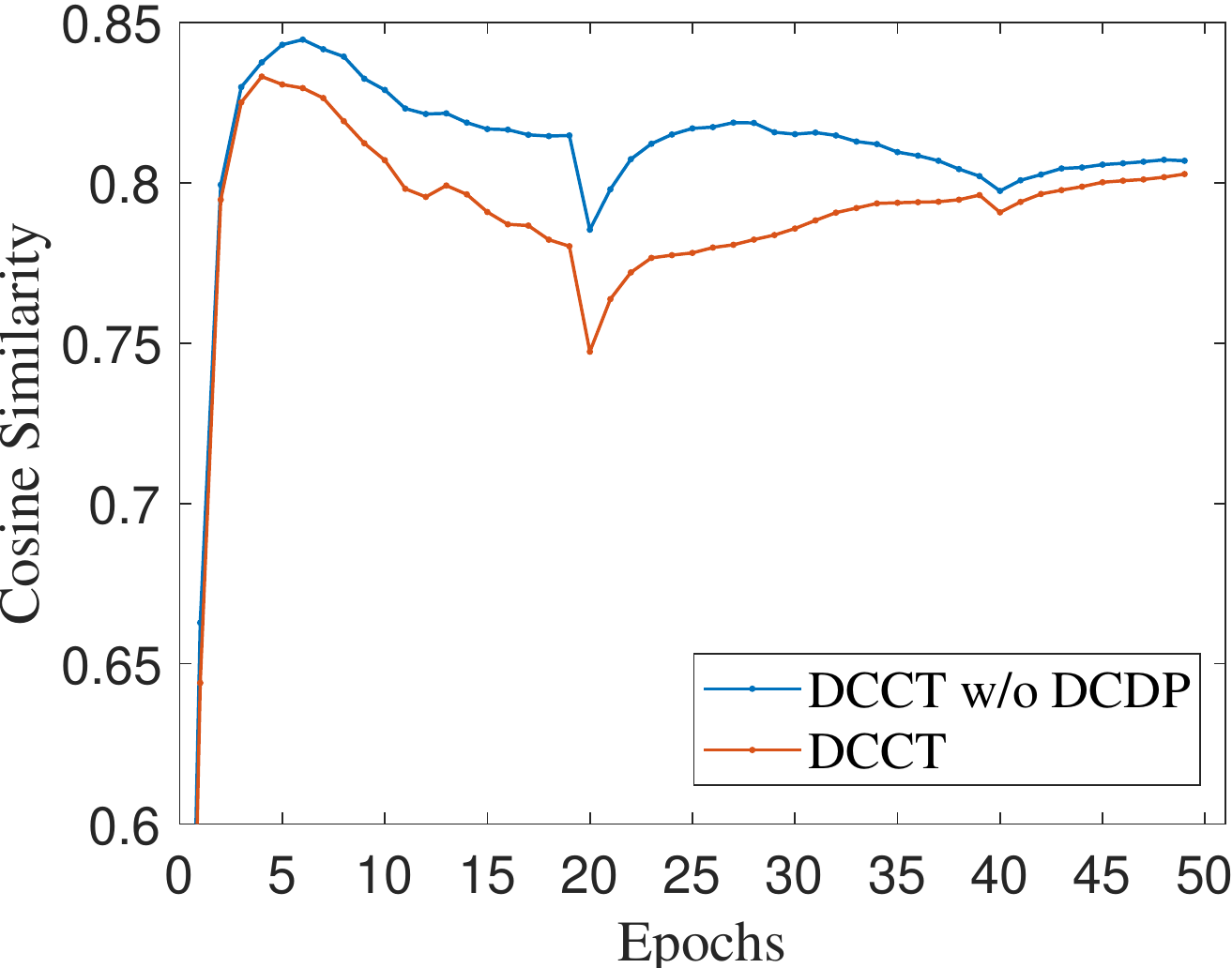}
		\label{DCDP_Figure_b}
	}
	
	\caption{(a) The cluster number of the two clusterings over different epochs with and without DCDP on Market-1501. (b) The average cosine similarity between the features of the two networks at different epochs with and without DCDP on Market-1501.}
	\label{DCDP_Figure}
\end{figure}	

\begin{figure}[!b]
	\centering
	\subfloat[]{
		\includegraphics[width=0.23\textwidth,trim=92 265 118 272,clip]{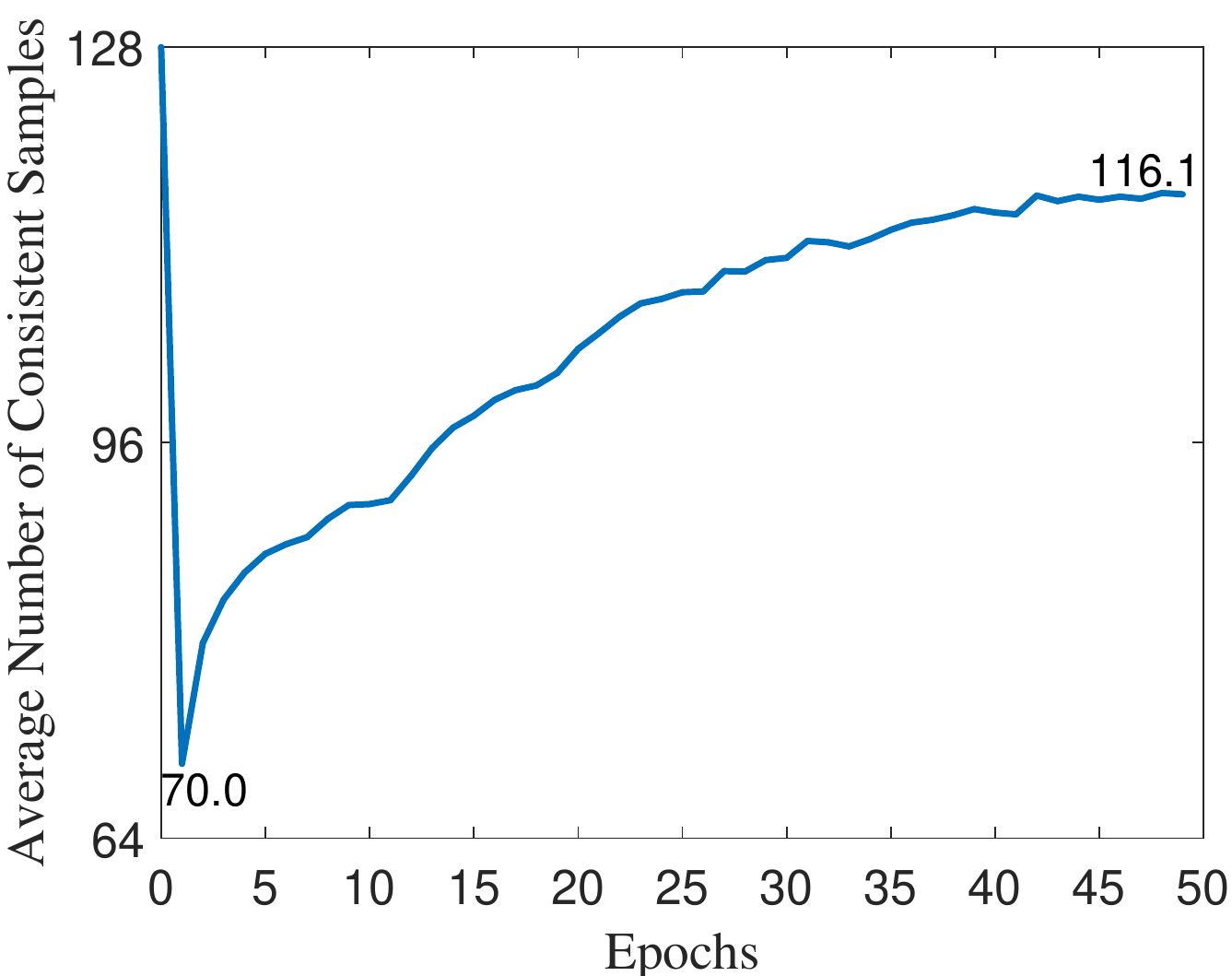}
		\label{CSM_Figure_a}
	}
	\subfloat[]{
		\includegraphics[width=0.23\textwidth,trim=92 265 118 272,clip]{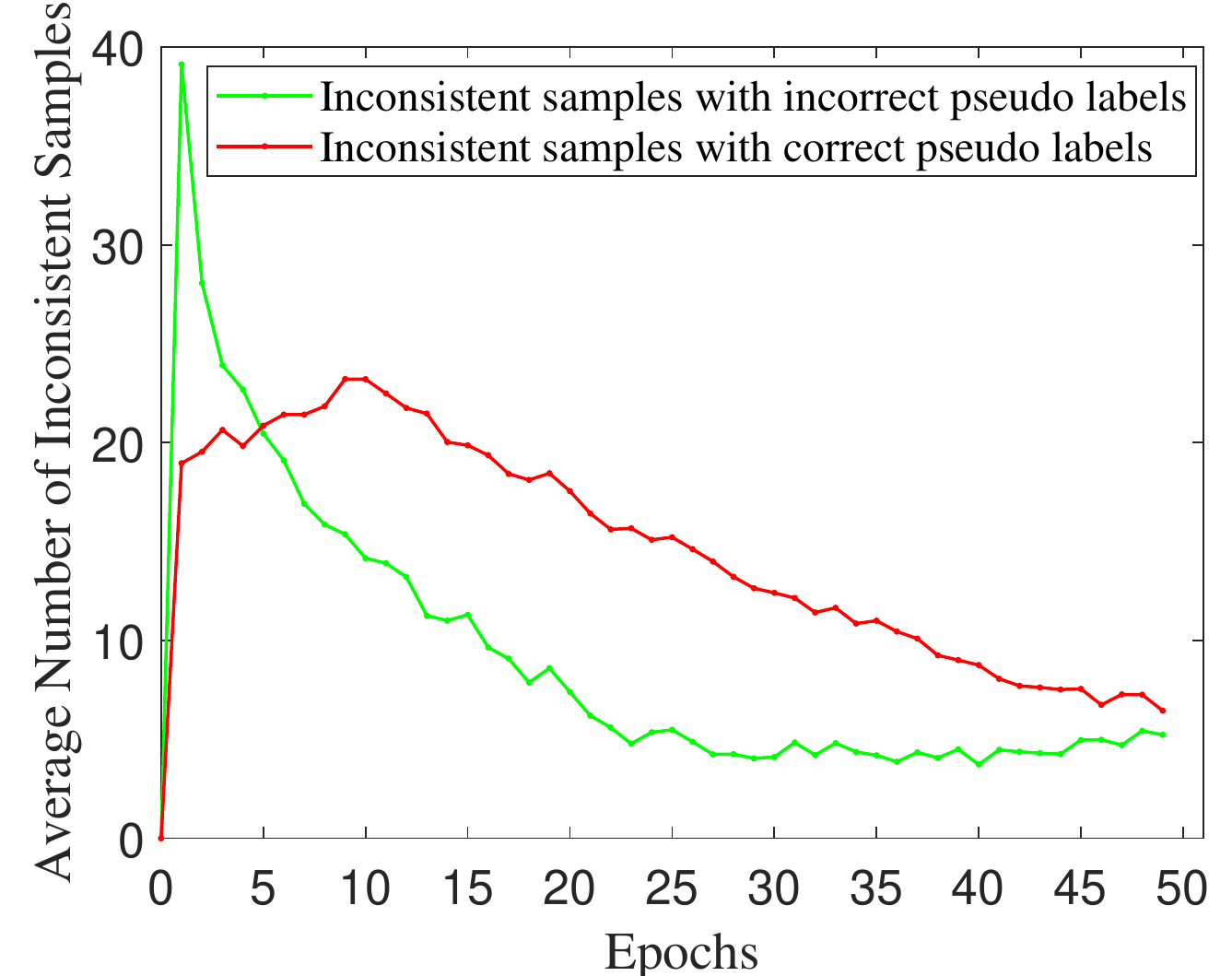}
		\label{CSM_Figure_b}
	}
	\caption{(a) The average number of consistent samples selected in each mini-batch over different epochs on Market-1501. (b) The average number of inconsistent samples with correct and incorrect pseudo labels in each mini-batch over different epochs on Market-1501.} 
	\label{CSM_Figure}
\end{figure}

\subsection{Ablation Study}

In this part, we verify the effectiveness of our proposed dual clustering with dynamic parameters (DCDP) and consistent sample mining (CSM).

\begin{figure*}[ht]
	\centering	
	\includegraphics[width=0.99\textwidth,trim=5 206 270 10,clip]{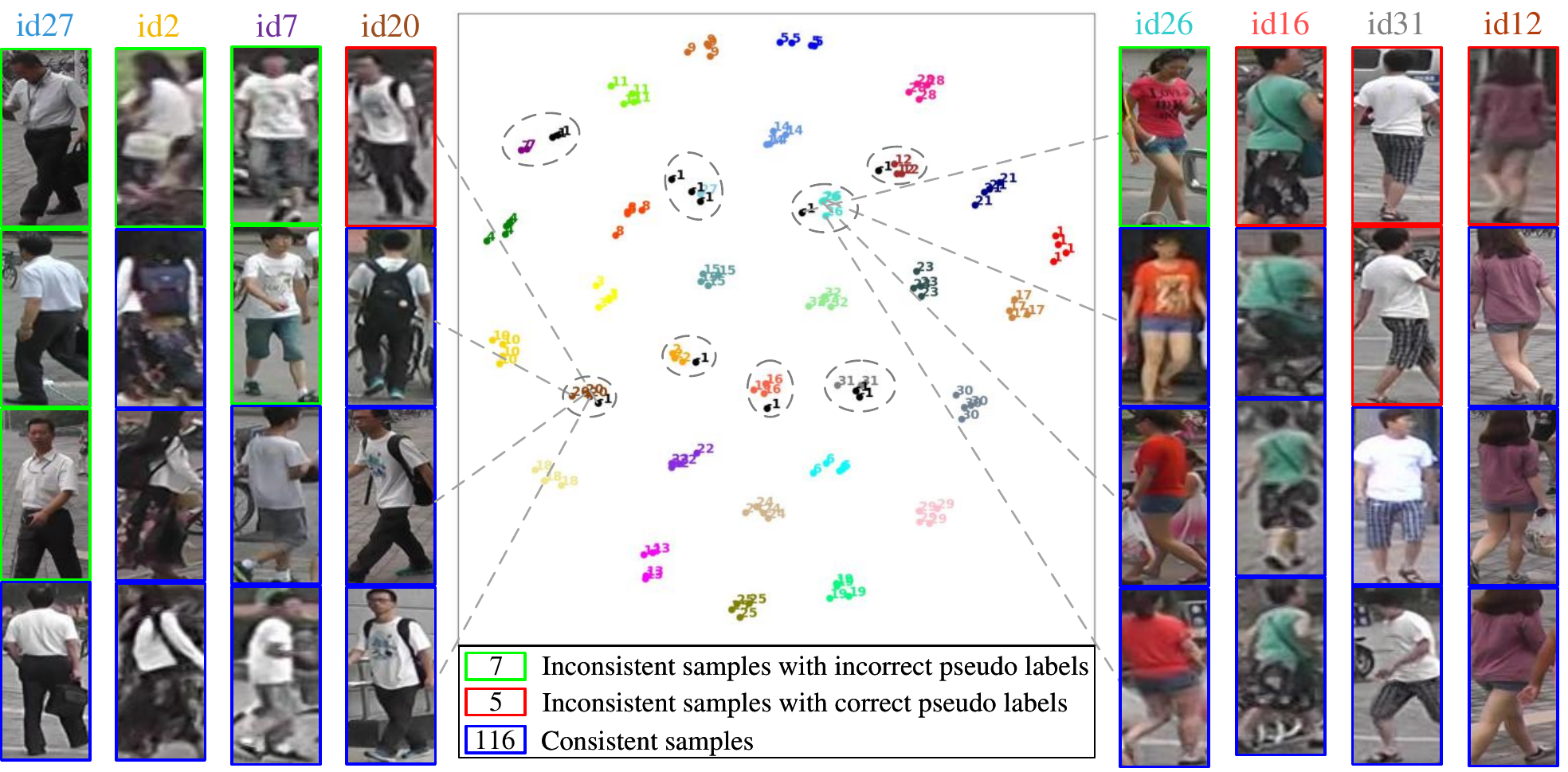}
	\caption{T-SNE \cite{2014tsne} visualization of the learned feature embeddings and their pseudo labels in the last iteration of the last epoch on the Market-1501 (32 identities, each identity with 4 images, total 128 images). The inconsistent samples (red and green boxes) mined by CSM are given a new pseudo label -1 and are not adopted for network training. Only consistent samples (blue boxes) are used for network training.} 
	\label{CSM_results}
\end{figure*}	

\subsubsection{Effectiveness of Dual Clustering with Dynamic Parameters}
The performances of using DCDP (denoted as ``DCCT'' and ``DCCT w/o CSM'') or not are shown in Table \ref{tab_ablation}. It can be observed that using the proposed DCDP can always obtain better performance on all three datasets. The reason is that the two networks are trained with pseudo labels obtained by different clusterings, which increases the differences and complementarity of the two networks. Therefore, the two networks can cope with different types of noises and improve the final performance. Meanwhile, the dynamically changing clustering parameters also enhance the network's generalization ability.

In Fig. \ref{DCDP_Figure}\subref{DCDP_Figure_a}, we compared the cluster number of the two clusterings with and without DCDP at different epochs on Market-1501. It can be observed that the difference in the number of clusters increases significantly when using DCDP, indicating that the two clusterings are more different. In Fig. \ref{DCDP_Figure}\subref{DCDP_Figure_b}, we compare the average cosine similarity between the features extracted by the two $Mean \ Nets$ with and without DCDP on Market-1501. Throughout the training process, the average cosine similarity with DCDP is always smaller than that without DCDP, indicating that DCDP increases the differences between the two networks. Since the learning rate is reduced to 1/10 of its previous value every 20 epochs, the cosine similarity has a noticeable drop every 20 epochs.

\subsubsection{Effectiveness of Consistent Sample Mining}
The performances of using CSM (denoted as ``DCCT'' and ``DCCT w/o DCDP'') or not are shown in Table \ref{tab_ablation}. It can be observed that using the proposed CSM can always obtain better performance on all three datasets. 

Fig. \ref{CSM_results} shows the CSM results of the last iteration in the last epoch on the Market-1501. It can be seen that some of the inconsistent samples mined by CSM have wrong pseudo labels (green boxes). Although some samples with correct pseudo labels may be considered inconsistent by CSM (red boxes), the experiment results in Table \ref{tab_ablation} show that training with wrong samples has a greater adverse effect than discarding a portion of the correct samples.

Fig. \ref{CSM_Figure}\subref{CSM_Figure_a} shows the average number of consistent samples selected in each mini-batch at different epochs on Market-1501. It can be observed that the number of selected consistent samples gradually increases as the training progresses. Furthermore, the average numbers of inconsistent samples with correct and incorrect pseudo labels are also illustrated in Fig. \ref{CSM_Figure}\subref{CSM_Figure_b}. We can see that in the early stage of training, the vast majority of inconsistent samples have incorrect pseudo labels. Dropping these inconsistent samples can effectively reduce the impact of noises on training. Although about half of the discarded samples have correct pseudo labels, the experimental results in Table \ref{tab_ablation} show that the negative impact of discarding some correct samples is smaller than the positive impact of discarding those incorrect samples.

\subsubsection{Qualitative Analysis of T-SNE Visualization}

\begin{figure}[b!]
	\centering	
	\includegraphics[width=0.49\textwidth,trim=8 262 375 5,clip]{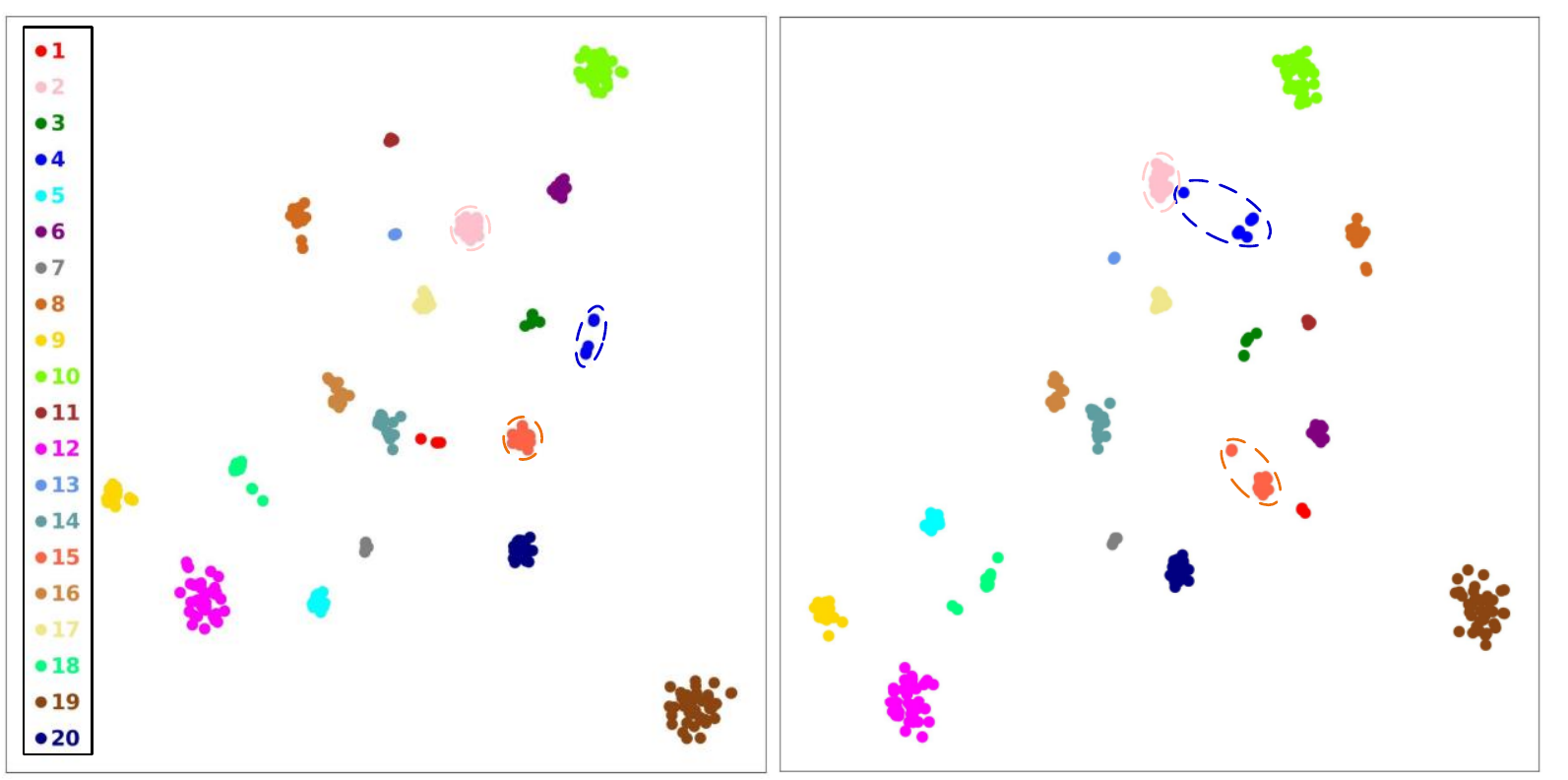}
	\caption{T-SNE \cite{2014tsne} visualization of 20 random identities on Market-1501 between DCCT (Left) and DCCT w/o (DCDP \& CSM) (Right). Different numbers and colors represent different identities.} 
	\label{tsne_identity}
\end{figure}	

To further illustrate that the proposed DCDP and CSM can improve the model's discriminative ability, we employ T-SNE to visualize the feature embeddings of person images with 20 random identities on Market-1501. As shown in Fig. \ref{tsne_identity}, after employing DCDP and CSM, the feature distribution of persons with the same identity is more compact. Furthermore, there is less mixing and overlap among person features with different identities.

\subsection{Parameter Analysis}
\label{Sec_parameter_analysis}

\begin{figure*}[!t]
	\centering
	
	\subfloat[The parameter analyses of maximum distance $\varepsilon$ on the three datasets.]{
		\includegraphics[width=0.31\textwidth,trim=100 265 125 270,clip]{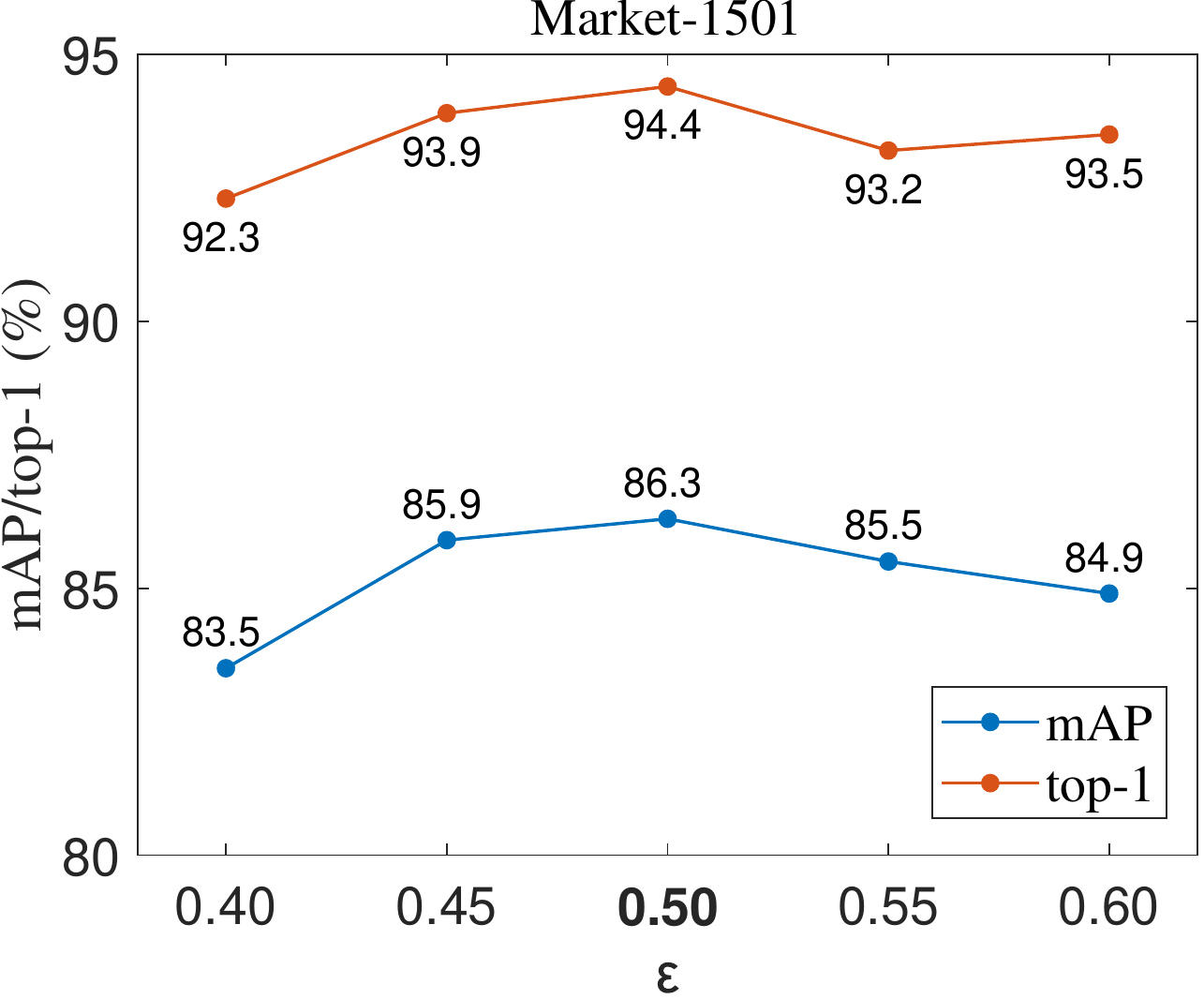}
		\hspace{3pt}
		\includegraphics[width=0.31\textwidth,trim=100 265 125 270,clip]{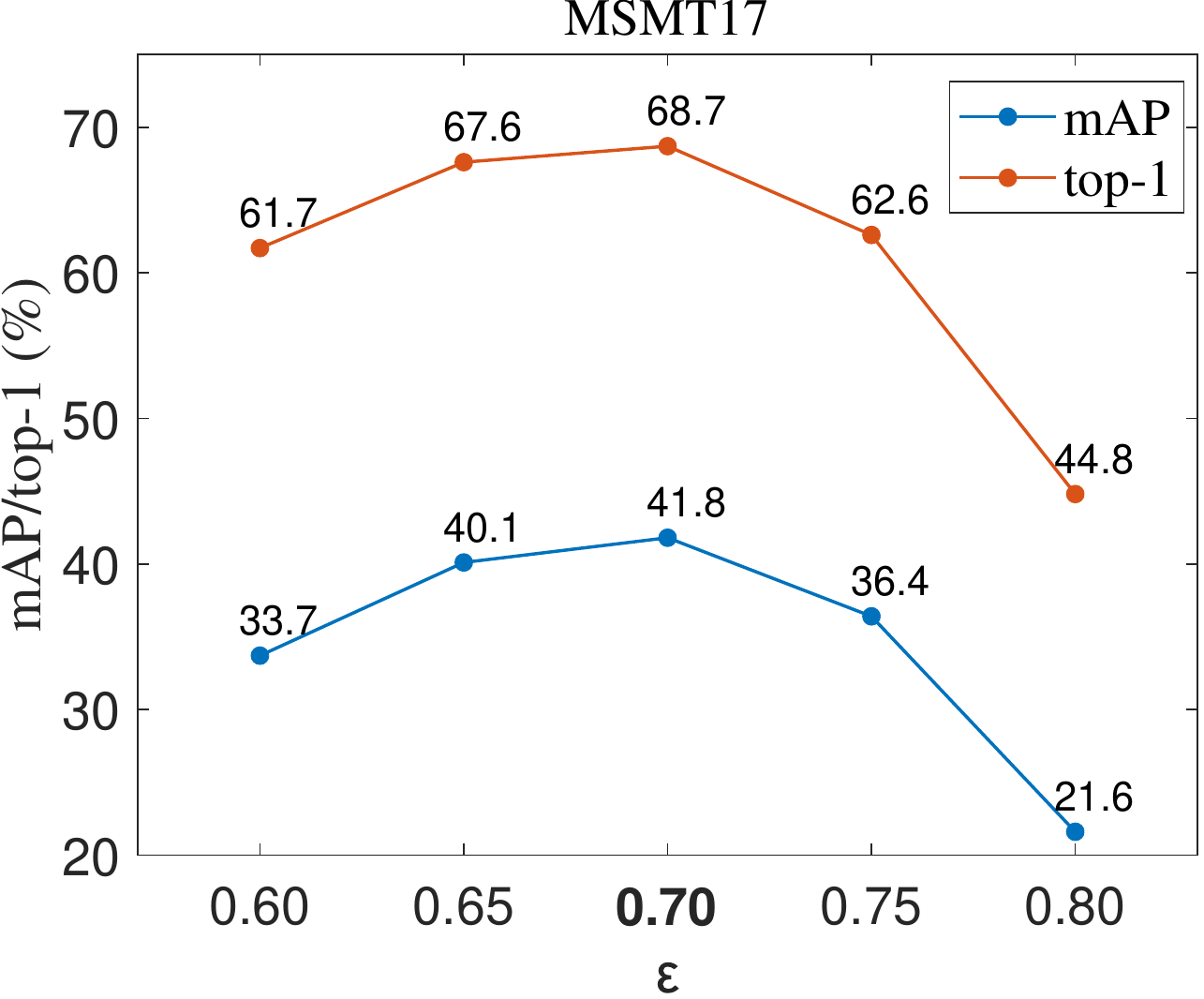}
		\hspace{3pt}
		\includegraphics[width=0.31\textwidth,trim=100 265 125 270,clip]{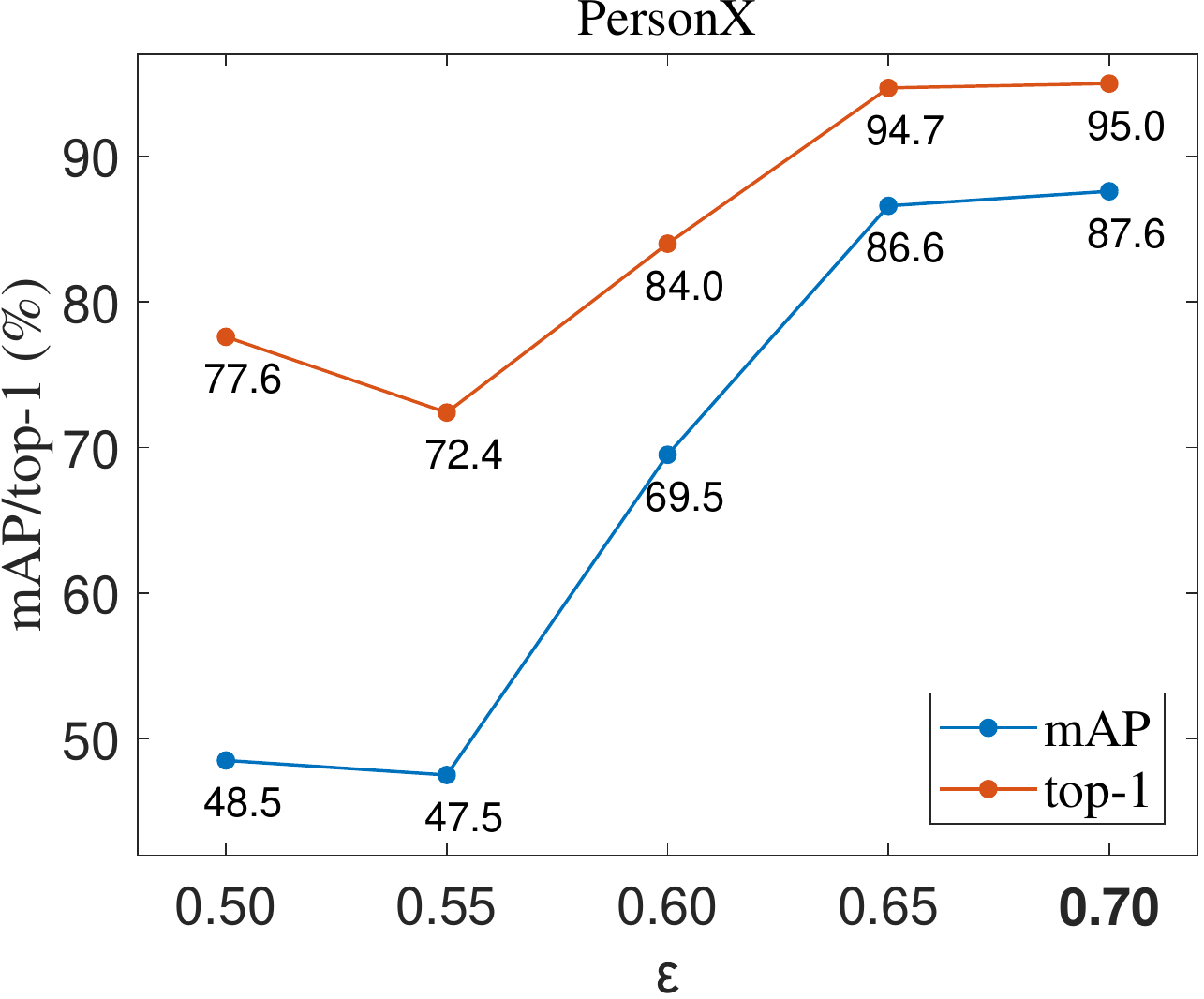}
		\label{Fig_Parameter_eps}
	}
	
	\subfloat[The parameter analyses of $\Delta{\varepsilon}$ in Eq. \ref{calculation_of_eps} on the three datasets.]{
		\includegraphics[width=0.31\textwidth,trim=100 265 125 270,clip]{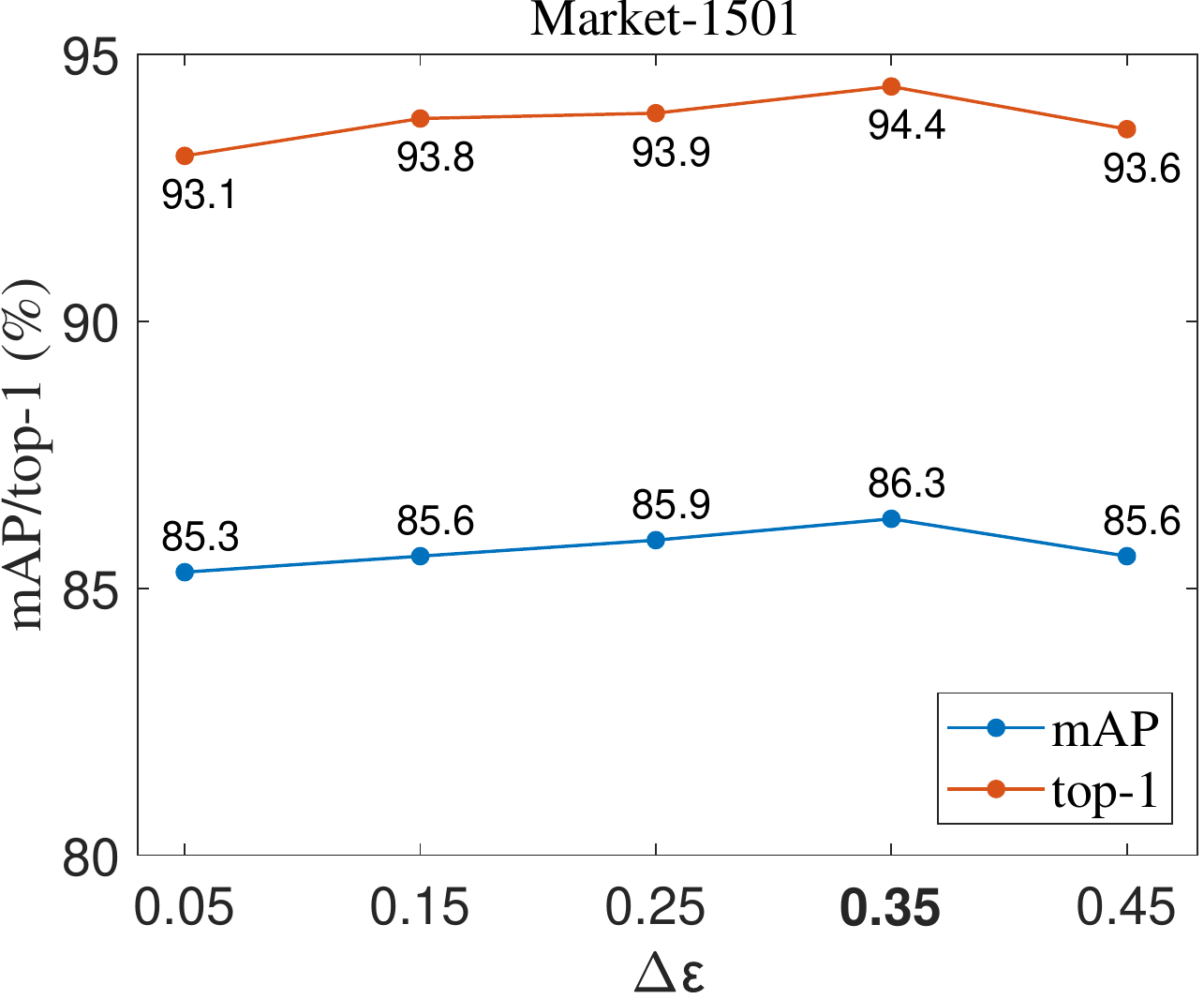}
		\hspace{3pt}
		\includegraphics[width=0.31\textwidth,trim=100 265 125 270,clip]{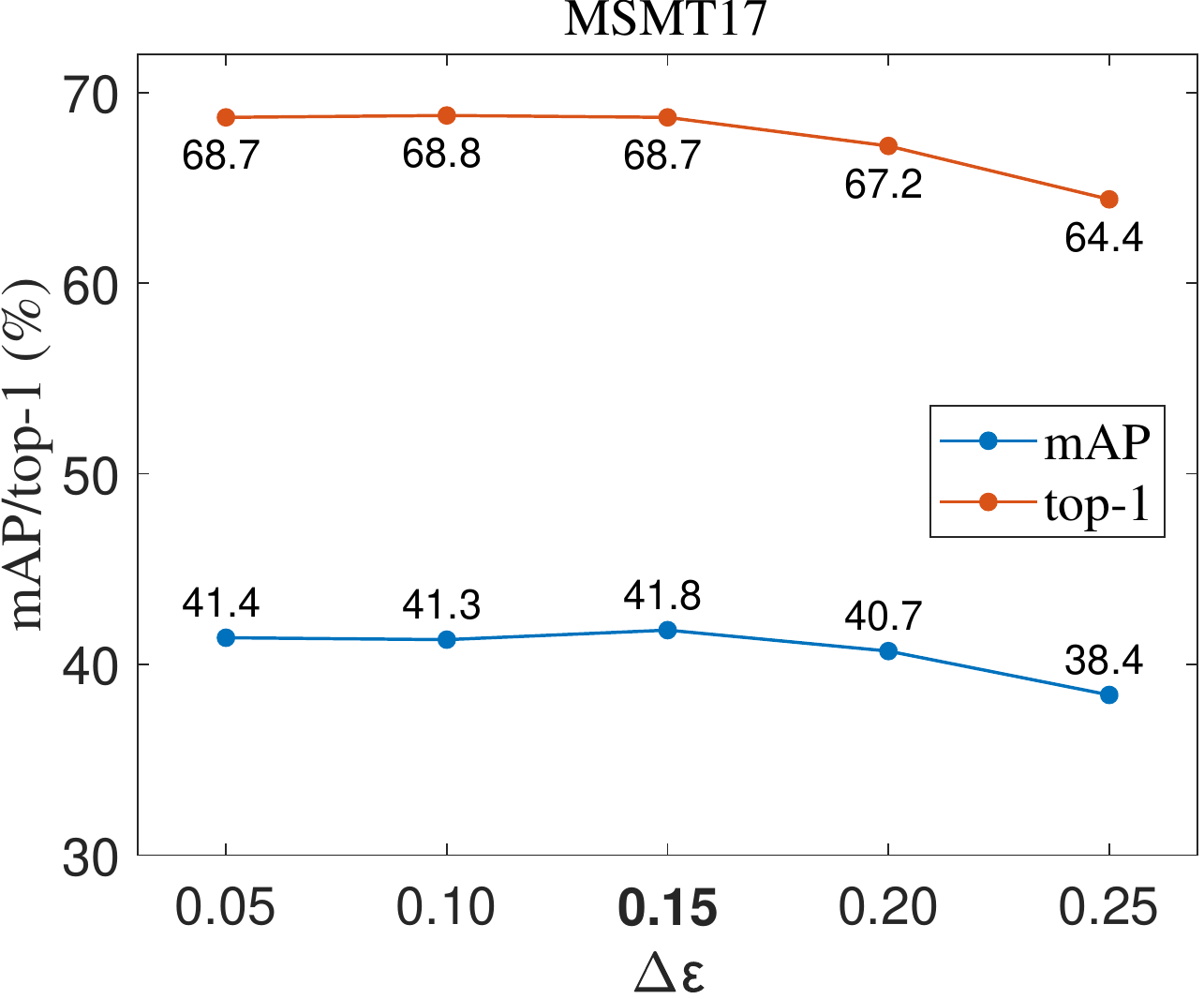}
		\hspace{3pt}
		\includegraphics[width=0.31\textwidth,trim=100 265 125 270,clip]{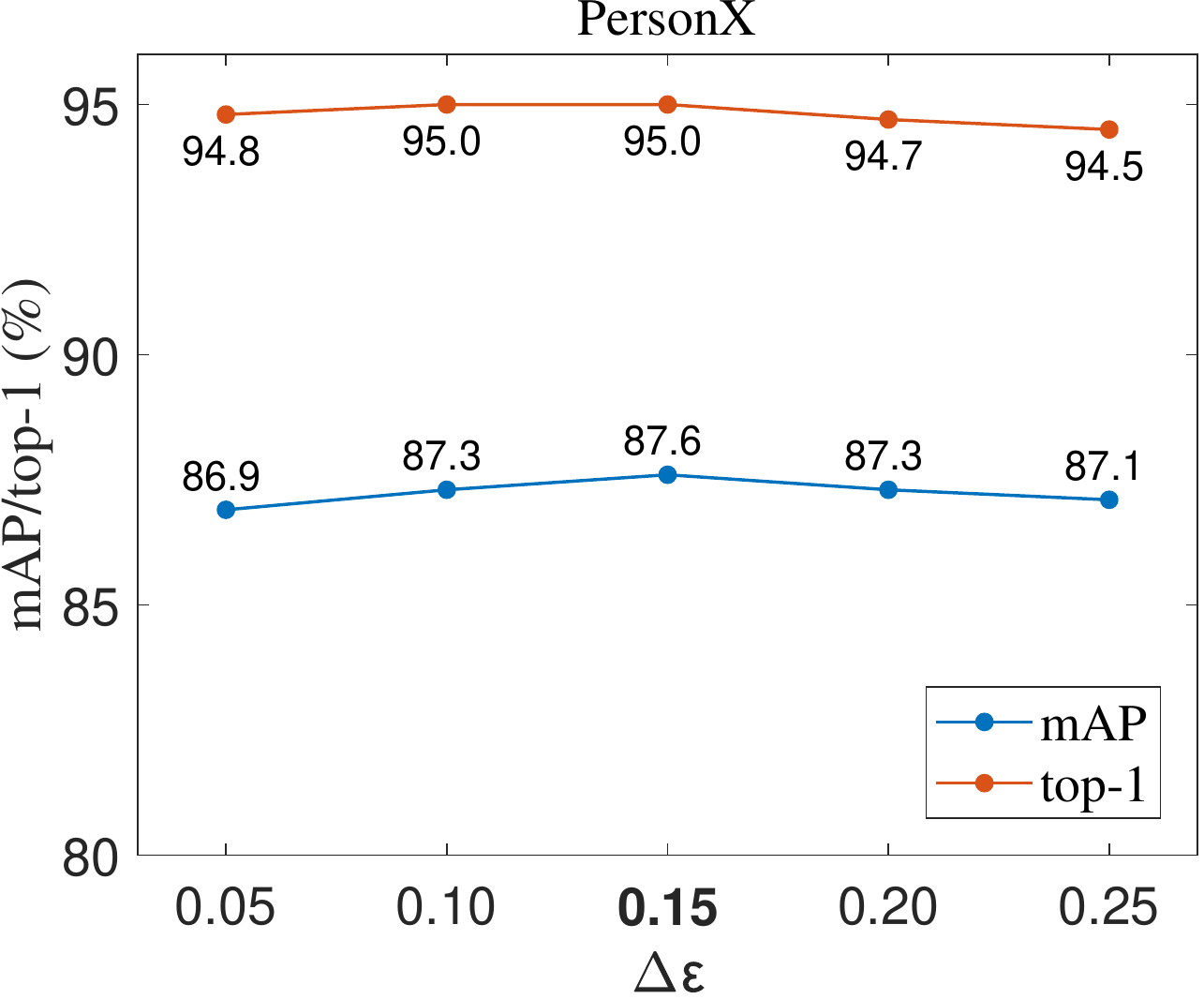}
		\label{Fig_Parameter_DCDP}
	}
	
	\subfloat[The parameter analyses of $\gamma$ in Sec. \ref{Detail_CSM} on the three datasets.]{
		\includegraphics[width=0.31\textwidth,trim=100 265 125 270,clip]{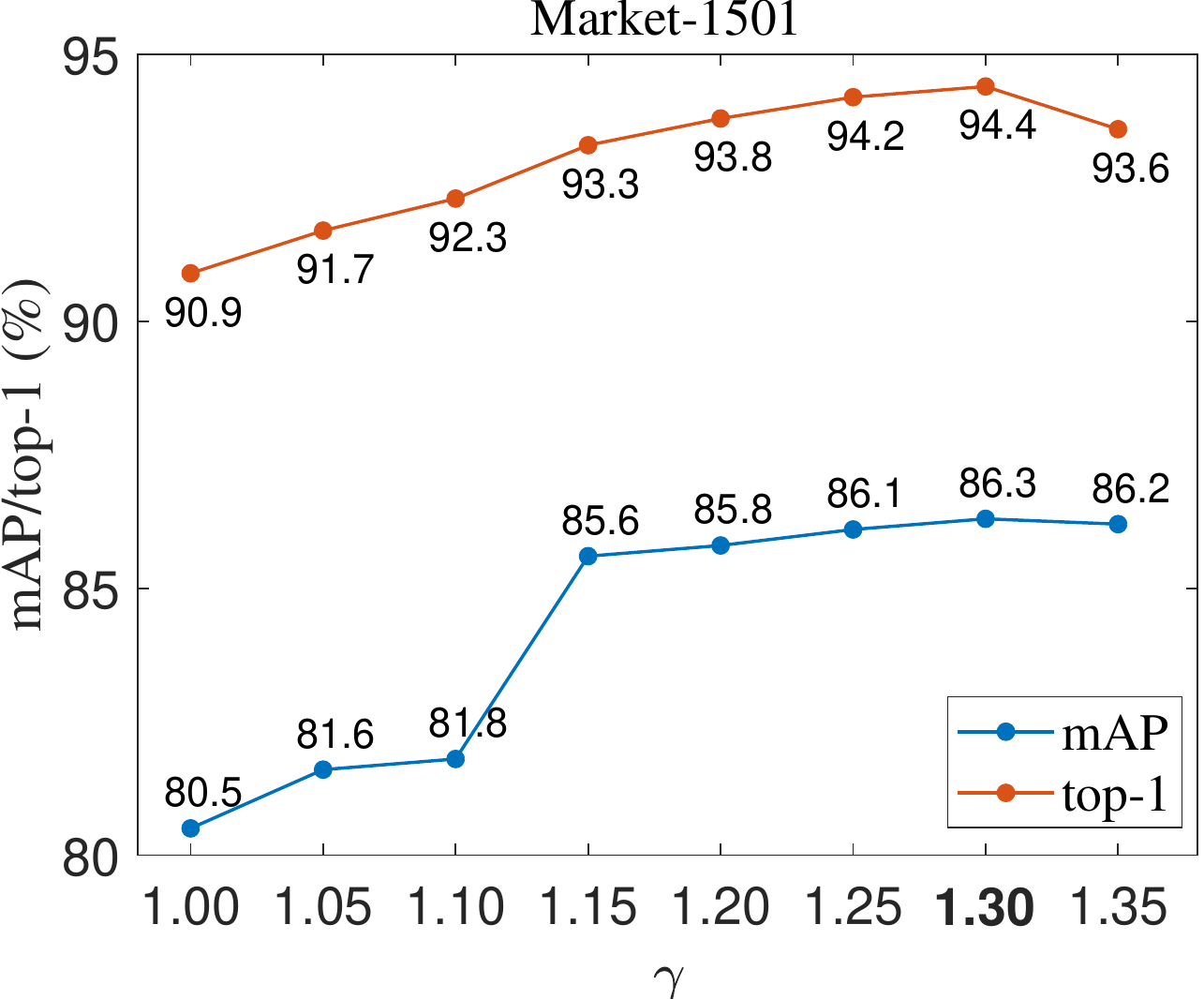}
		\hspace{3pt}
		\includegraphics[width=0.31\textwidth,trim=100 265 125 270,clip]{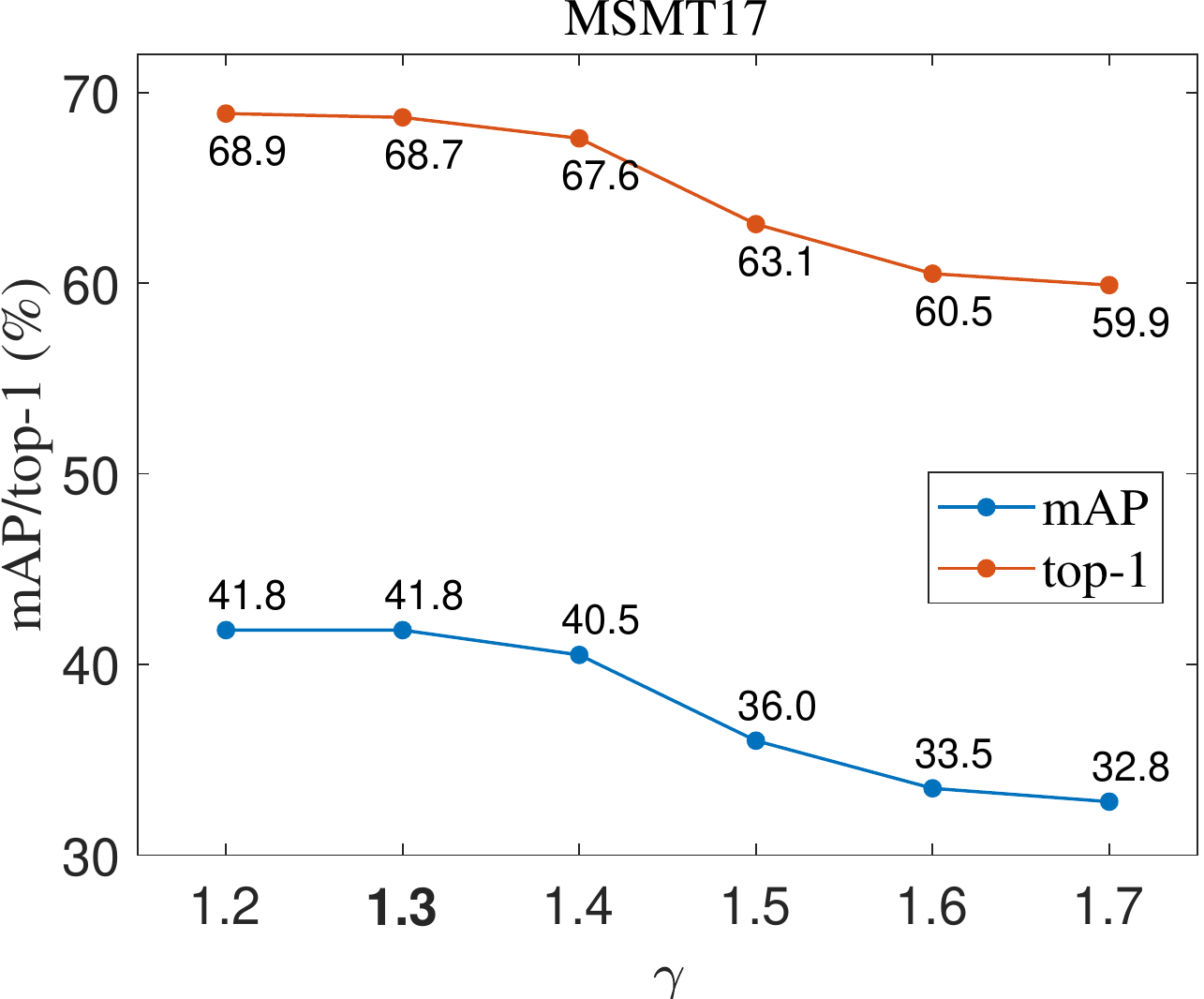}	
		\hspace{3pt}
		\includegraphics[width=0.31\textwidth,trim=100 265 125 270,clip]{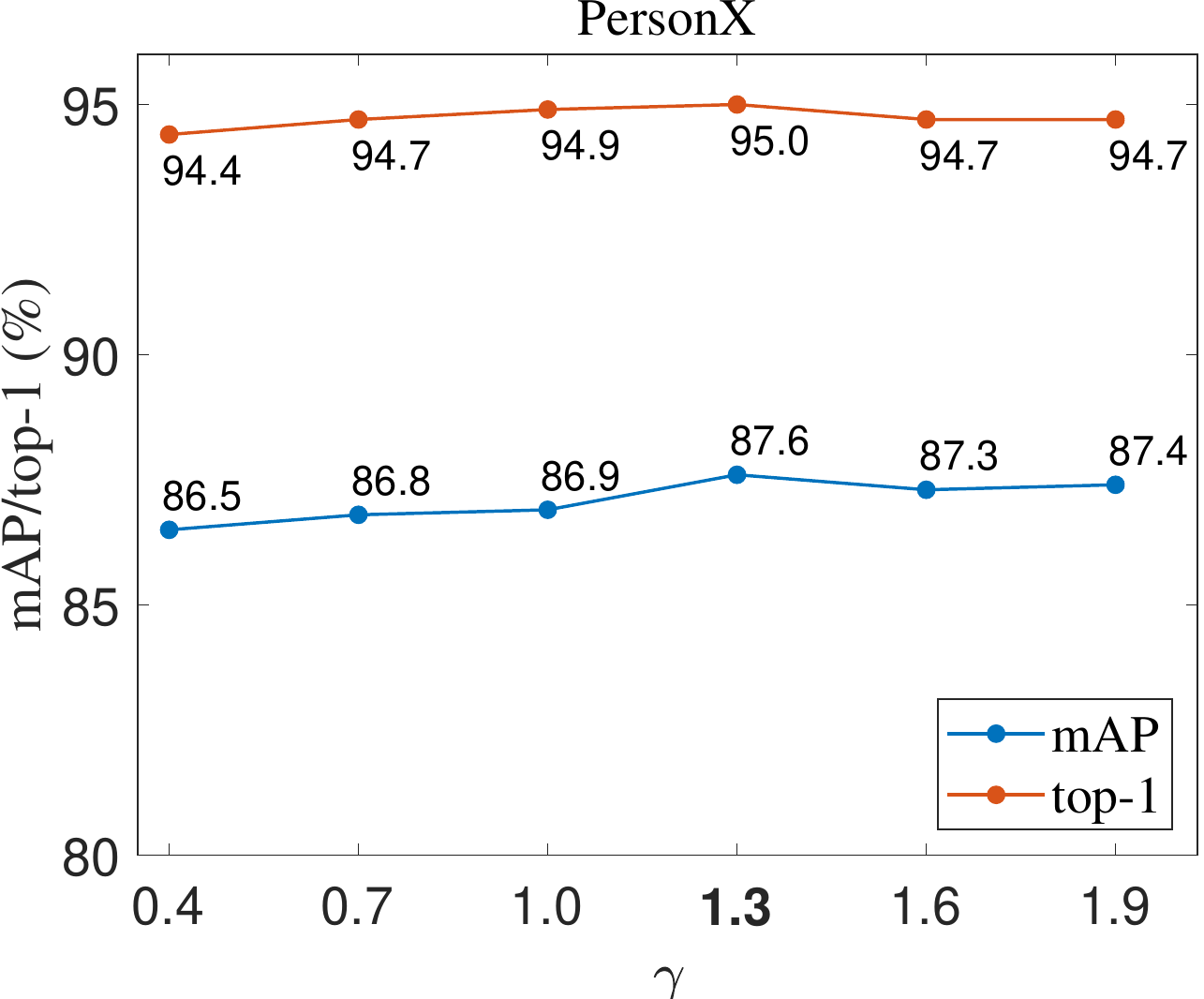}
		\label{Fig_Parameter_CSM}
	}
	
	\caption{The parameter analyses on the three datasets. The optimal parameters are \textbf{bold}.} 
	\label{Fig_Parameter_Analyses}
\end{figure*}

\subsubsection{Maximum Distance for DBSCAN}
Due to distribution differences on various datasets, many state-of-the-art methods use inconsistent clustering hyper-parameters on different datasets \cite{2022PPLR,2022ISE,2021ICE,2021IICS}. In DBSCAN \cite{1996DBSCAN}, hyper-parameter $\varepsilon$ represents the maximum distance between two samples. DBSCAN with a smaller $\varepsilon$ tends to group persons with the same identity into different clusters. Conversely, DBSCAN with a larger $\varepsilon$ tends to group persons with different identities into the same cluster. Both too large and too small $\varepsilon$ degrade the clustering quality and hinder the network training.

Fig. \ref{Fig_Parameter_Analyses}\subref{Fig_Parameter_eps} shows the sensitivity of the performance of DCCT to $\varepsilon$ on three datasets. The optimal $\varepsilon$ on each dataset is \textbf{bold}. It can be observed that the best value of $\varepsilon$ on the PersonX dataset is 0.7. When $\varepsilon$ is increased to 0.75 on PersonX, the network cannot converge due to too few clusters. The optimal $\varepsilon$ on Market-1501 and MSMT17 is 0.5 and 0.7, respectively. However, different methods may have different optimal $\varepsilon$ on the same dataset. State-of-the-art unsupervised person Re-ID method ISE \cite{2022ISE} sets $\varepsilon$ to 0.4 on Market-1501 and 0.7 on MSMT17. And SOTA method PPLR \cite{2022PPLR} sets $\varepsilon$ to 0.6 on Market-1501 and 0.7 on MSMT17. Both the above methods set $\varepsilon$ on Market-1501 smaller than that on MSMT17, which is consistent with our experimental results. 

\subsubsection{Hyper-parameters Introduced in Our Method }
Our method introduces hyper-parameters $\Delta{\varepsilon}$ (see Eq. \ref{calculation_of_eps}) and $\gamma$ (see Sec. \ref{Detail_CSM}) related to clustering. As mentioned above, the best clustering hyper-parameters are usually inconsistent on different datasets due to distribution differences. Therefore, we tune the hyper-parameters on the three datasets.

The optimal hyper-parameters on each dataset are \textbf{bold} in Fig. \ref{Fig_Parameter_Analyses}\subref{Fig_Parameter_DCDP}. It can be observed that both too large and too small $\Delta{\varepsilon}$ lead to performance degradation. Too small $\Delta{\varepsilon}$ provides little difference between the two clusterings, leading to a finite improvement in the difference and complementarity of the network. While too large $\Delta{\varepsilon}$ may result in a poor clustering parameter $\varepsilon$, which reduces the clustering quality and hinders the network's training.

As shown in Fig. \ref{Fig_Parameter_Analyses}\subref{Fig_Parameter_CSM}, the best value for $\gamma$ on all three datasets is 1.3. Experiments on Market-1501 show that network performance is degraded due to too small $\gamma$. The reason is that CSM starts too late, and many noise samples are used for network training. The results on MSMT17 show that too large $\gamma$ also hinders network training. The reason is that prematurely mining consistent samples on low-quality clustering results in too few samples available for network training. The network performance on PersonX is not sensitive to $\gamma$.

\subsection{The effectiveness of DCDP on other clustering algorithms}
\label{DCDP_on_other}

This experiment aims to demonstrate that the proposed DCDP is applicable for not only DBSCAN but also InfoMap \cite{2008infomap} and $k$-means \cite{1982kmeans}. 

\subsubsection{InfoMap}
\begin{figure}[!b]
	\centering
	
	\subfloat[The optimal $\psi$ for InfoMap (DCCT w/o DCDP).]{
		\includegraphics[width=0.22\textwidth,trim=95 265 125 270,clip]{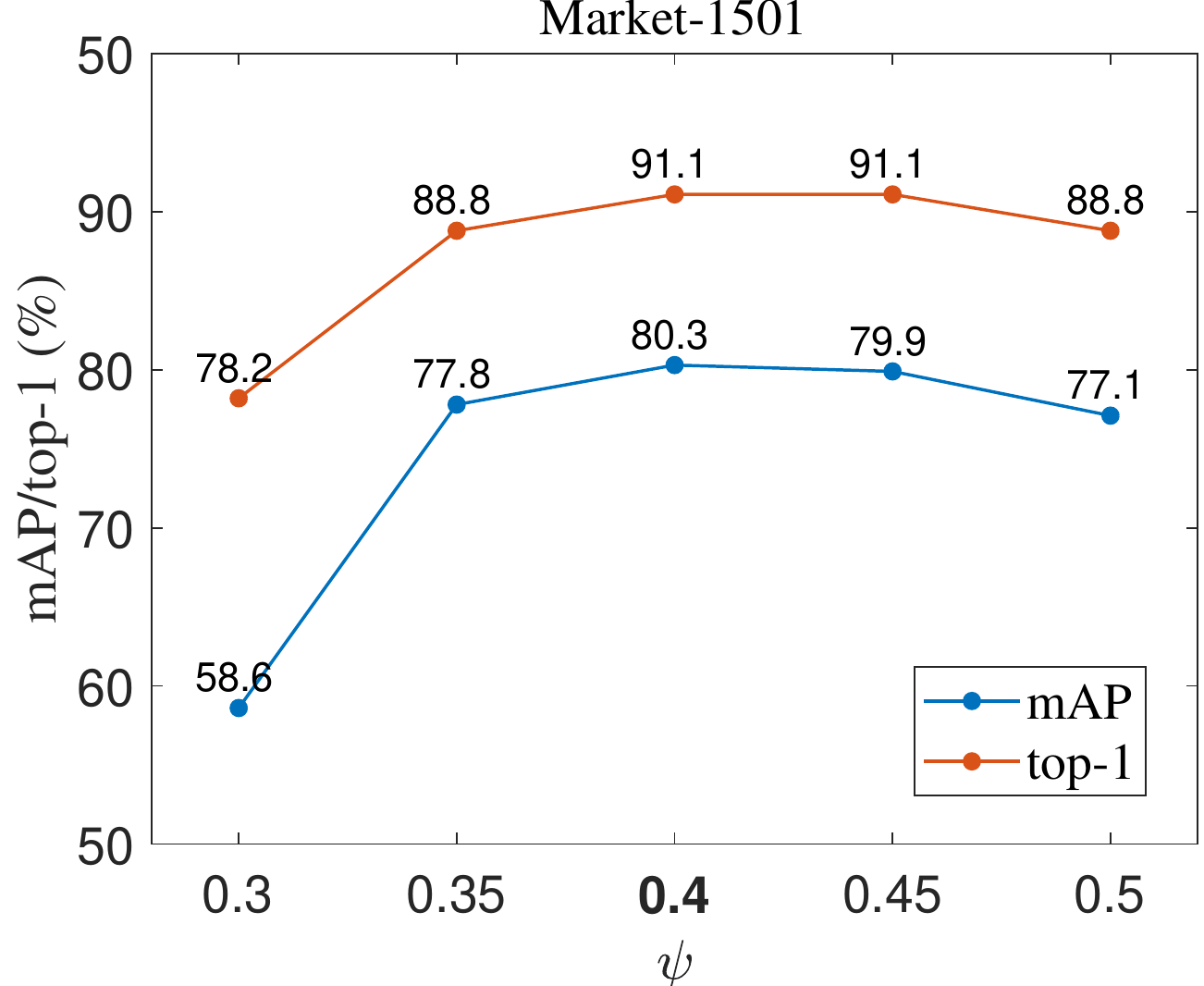}
		\label{Fig_infomap_a}
	}
	\hspace{2pt}
	\subfloat[The optimal $\Delta{\psi}$ for InfoMap (DCCT).]{
		\includegraphics[width=0.22\textwidth,trim=95 265 125 270,clip]{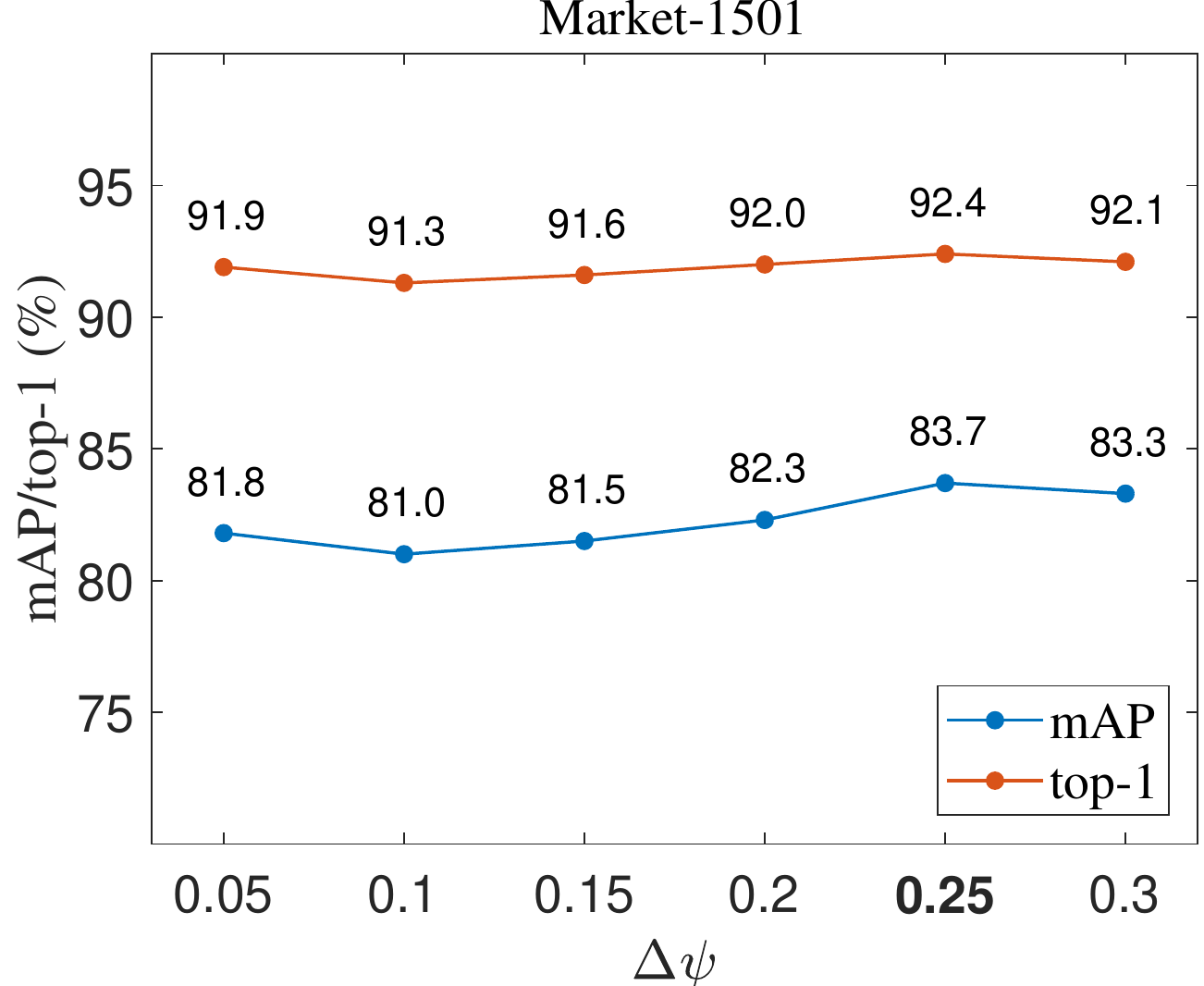}
		\label{Fig_infomap_b}
	}
	
	\subfloat[The optimal cluster number $k$ for $k$-means (DCCT w/o DCDP).]{
		\includegraphics[width=0.22\textwidth,trim=95 265 125 270,clip]{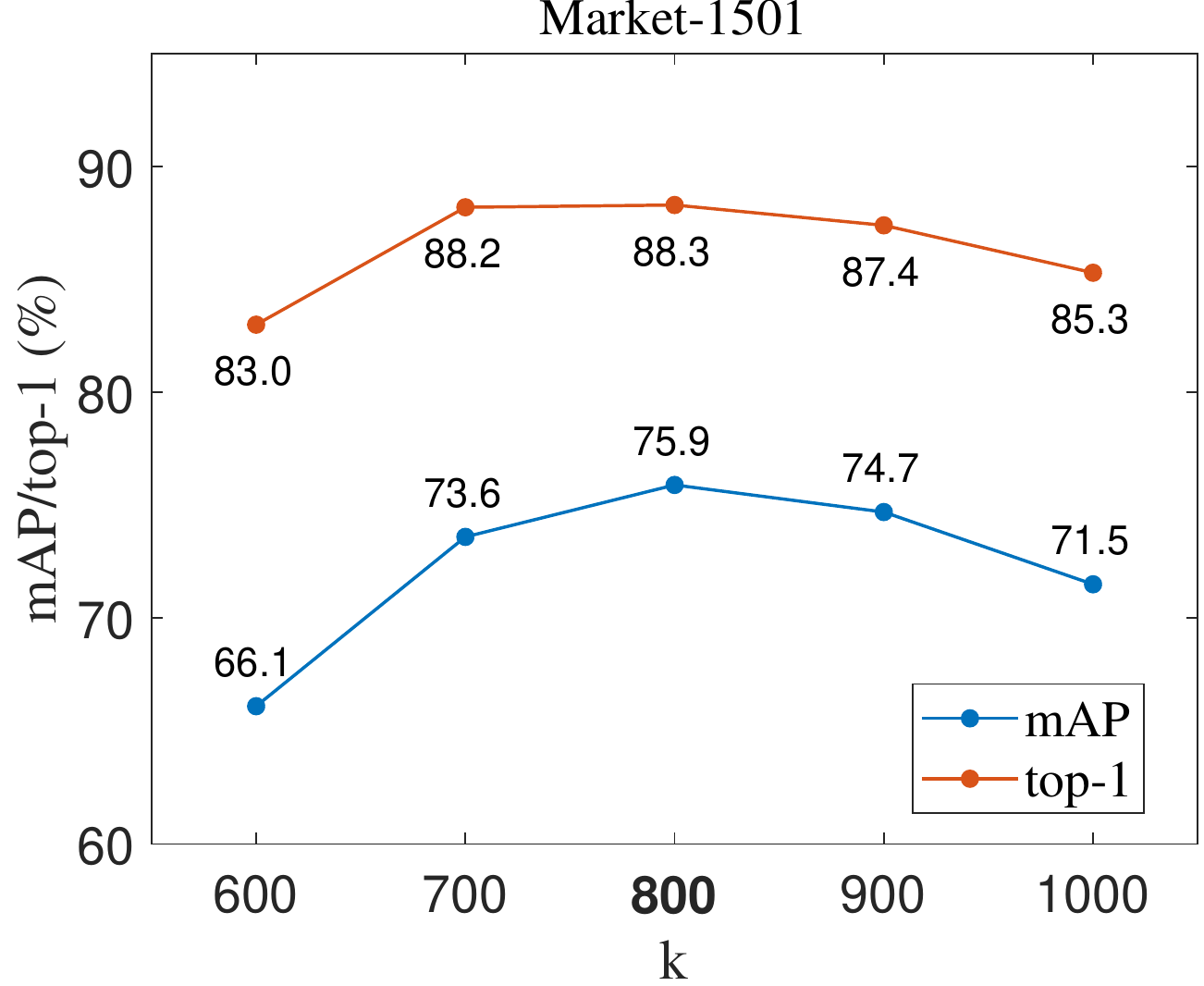}
		\label{Fig_kmeans_a}
	}
	\hspace{2pt}
	\subfloat[The optimal $\Delta{k}$ for $k$-means (DCCT).]{
		\includegraphics[width=0.22\textwidth,trim=95 265 125 270,clip]{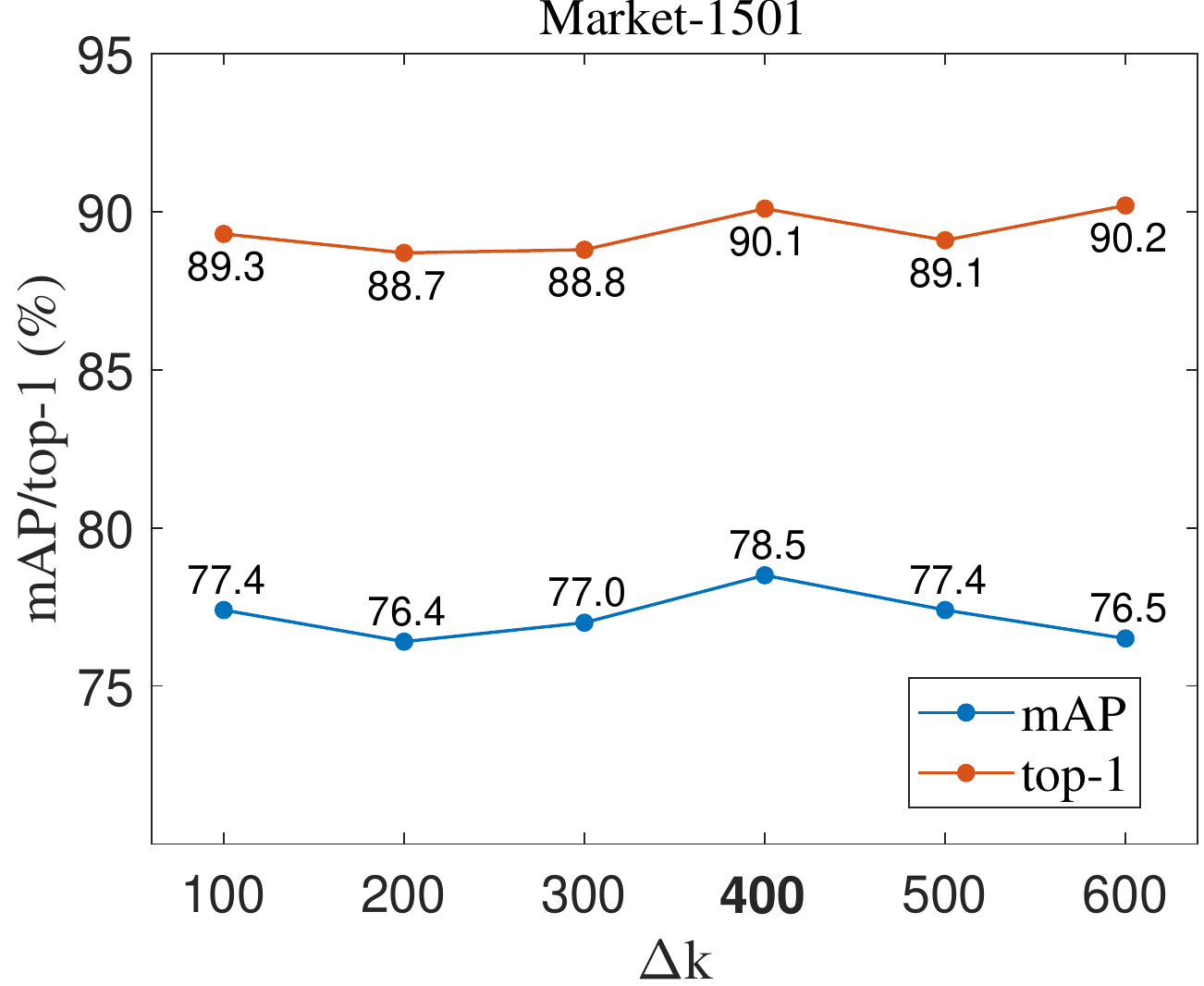}
		\label{Fig_kmeans_b}
	}
	
	\caption{The effectiveness of DCDP on other clustering algorithms.} 
	\label{Fig_DCDP_on_other}
\end{figure}

To use InfoMap for clustering, we need to convert all samples into a directed graph, where nodes are samples. Let $D(i,j)$ represents the distance between any two samples, we link the two nodes when $D(i,j)$ is less than maximum distance $\psi$. And the weight of the link is represented as $1- D(i,j)$. We adopt $\psi$ as the dynamic parameter in DCDP. Given the initial value $\psi$ and the increment size $\Delta{\psi}$ of the maximum distance, the maximum distance $\psi_1$ and $\psi_2$ of $clustering_1$ and $clustering_2$ will vary in the range of [$\psi - \Delta{\psi}$, $\psi + \Delta{\psi}$]. Then the $\psi_1^i$ and $\psi_2^i$ at the $i$-th epoch can be obtained from Eq. \ref{calculation_of_eps}. (Replace $\varepsilon$ with $\psi$ in Eq. \ref{calculation_of_eps}.) As shown in Fig. \ref{Fig_DCDP_on_other}\subref{Fig_infomap_a}, the best $\psi$ on Market-1501 is 0.4. After tuning $\Delta{\psi}$ in Fig. \ref{Fig_DCDP_on_other}\subref{Fig_infomap_b}, it can be observed that better performance can be obtained with DCDP. When $\Delta{\psi}$ is set to the optimal value of 0.25, the mAP and top-1 are improved by $3.4\%$ and $1.3\%$, respectively.

\subsubsection{$k$-means}The cluster number $k$ is the most crucial parameter in $k$-means, which is adopted as the dynamic parameter in DCDP. Given the initial value $k$ and the increment size $\Delta{k}$ of the cluster number, the cluster number $k_1$ and $k_2$ of $clustering_1$ and $clustering_2$ will vary in the range of [$k - \Delta{k}$, $k + \Delta{k}$]. Then the $k_1^i$ and $k_2^i$ at the $i$-th epoch can be obtained from Eq. \ref{calculation_of_eps}. (Replace $\varepsilon$ with $k$ in Eq. \ref{calculation_of_eps}, and round up the result to the nearest integer.) As shown in Fig. \ref{Fig_DCDP_on_other}\subref{Fig_kmeans_a}, the best cluster number $k$ on Market-1501 is 800. After tuning $\Delta{k}$ in Fig. \ref{Fig_DCDP_on_other}\subref{Fig_kmeans_b}, it can be observed that better performance can be obtained with DCDP, which is consistent with the conclusion on DBSCAN and InfoMap. When $\Delta{k}$ is set to the optimal value of 400, DCDP brings $+2.6\%$/$+1.8\%$ mAP/top-1 improvements.

\section{Conclusion}
This paper proposes a novel Dual Clustering Co-teaching (DCCT) framework to deal with noisy pseudo labels in unsupervised person Re-ID tasks. Unlike the previous peer-teaching methods utilizing a set of noisy pseudo labels to train the two networks, we propose dual clustering with dynamic parameters (DCDP) to generate two sets of pseudo labels for network training, which can increase the two networks' differences and complementarity, so that our method is more robust to the noisy pseudo labels. Furthermore, we also propose consistent sample mining (CSM) to find the samples with unchanged pseudo labels during training and remove potential noisy samples. Extensive experimental results on various person Re-ID datasets demonstrate that our method outperforms the prior state-of-the-art unsupervised methods.

\bibliographystyle{IEEEtran}
\bibliography{DCCT}

\newpage

\vfill

\end{document}